\documentclass[final,1p,times]{elsarticle}

\usepackage[colorlinks=true, allcolors=blue]{hyperref}
\usepackage[toc,page]{appendix}
\usepackage{subfigure}
\usepackage{multirow}
\usepackage[symbol]{footmisc}
\usepackage[normalem]{ulem}
\usepackage{tabu}

\usepackage{amsmath}
\usepackage[colorinlistoftodos]{todonotes}

\usepackage{graphicx}
\usepackage{amssymb}

\journal{Computer Methods and Programs in Biomedicine}

\begin{document}

\begin{frontmatter}


\title{An Empirical Evaluation of Deep Learning for ICD-9 Code Assignment using MIMIC-III Clinical Notes}



\author{Jinmiao Huang\corref{cor}}
\ead{vichuang@gatech.edu}
\address{Georgia Institute of Technology, Atlanta, Georgia, USA}

\author{Cesar Osorio\corref{cor1}}
\ead{cesar.osorio@gatech.edu}
\address{Georgia Institute of Technology, Atlanta, Georgia, USA}

\author{Luke Wicent Sy\corref{cor1}}
\ead{sylukewicent@gmail.com}
\address{Georgia Institute of Technology, Atlanta, Georgia, USA}

\cortext[cor]{corresponding author}

\begin{abstract}
\textbf{Background and Objective:} Code assignment is of paramount importance in many levels in modern hospitals, from ensuring accurate billing process to creating a valid record of patient care history. However, the coding process is tedious and subjective, and it requires medical coders with extensive training. This study aims to evaluate the performance of deep-learning-based systems to automatically map clinical notes to ICD-9 medical codes. \textbf{Methods:} The evaluations of this research are focused on end-to-end learning methods without manually defined rules. Traditional machine learning algorithms, as well as state-of-the-art deep learning methods such as Recurrent Neural Networks and Convolution Neural Networks, were applied to the Medical Information Mart for Intensive Care (MIMIC-III) dataset. An extensive number of experiments was applied to different settings of the tested algorithm. \textbf{Results:} Findings showed that the deep learning-based methods outperformed other conventional machine learning methods. From our assessment, the best models could predict the top 10 ICD-9 codes with 0.6957 $F_1$ and 0.8967 accuracy and could estimate the top 10 ICD-9 categories with 0.7233 $F_1$ and 0.8588 accuracy. Our implementation also outperformed existing work under certain evaluation metrics. \textbf{Conclusion:} A set of standard metrics was utilized in assessing the performance of ICD-9 code assignment on MIMIC-III dataset. All the developed evaluation tools and resources are available \href{https://github.com/lsy3/clinical-notes-diagnosis-dl-nlp}{online}, which can be used as a baseline for further research.
\end{abstract}


\end{frontmatter}



\section{Introduction}
\label{sec:intro}

Electronic health record (EHR) data include a variety of patient clinical information such as medical history, vital signs, lab test results, and clinical notes. Such data can help in building a continuous flow of information between doctors and patients. More so, systematic reviews have shown that clinical care quality can be improved considerably using predictive analysis based on EHR data \cite{black2011impact}. 

EHR data contain both structured (e.g., blood pressure) and unstructured data (e.g., doctor's observation). While many medical systems focus on structured biosignal features in EHRs to build the clinical decision making systems \cite{choi2016doctor},  more than 80\% of  health record data are unstructured text \cite{martin2014big}. For example, clinical notes contain information about patients' medical history and doctors' observations and comments regarding their interactions with patients. 


The systems evaluated in this paper assign ICD-9 codes from a patient's free-text EHR. These codes can be subsequently used in billing or creating a valid record of patient care history. Currently, the task of assigning diagnosis codes is carried out manually by medical coders. Also, the volume of medical records generated makes the manual classification of diagnoses a labor-intensive process, thus resulting in a significant backlog of work. Automating ICD-9 code assignment will not only make the clinical process more efficient, but it will also take note of all EHRs and provide support to expedite some levels of semantic analysis which can help clinicians diagnose and improve the medical care systems effectively. Over the past two decades, researchers have explored machine learning methods to assign ICD-9 codes based on clinical notes, such as Support Vector Machine (SVM) \cite{ferrao2013using}, Naive Bayes \cite{pakhomov2006automating, medori2010machine}, and Boosting \cite{goldstein2007three}. Despite their research efforts, it is believed that the accuracy of this problem can be further improved, especially with the recent breakthrough in deep learning approaches. Deep learning techniques have shown a significant improvement in many Natural Language Processing (NLP) tasks such as language translation \cite{sutskever2014sequence}, natural language understanding \cite{collobert2011natural}, and sentiment analysis \cite{socher2011semi}. What's more, deep learning models can often be trained end-to-end without any domain-specific and hand-designed feature engineering, a process that is often tedious.

Furthermore, there is a lack of a baseline for the community to reliably assess different algorithms on benchmark datasets. Because of this challenge, this paper will focus on evaluating the performance of state-of-the-art deep neural networks to diagnose learning systems on a widely-used and publicly available dataset. Our results will be compared with several traditional classification systems, including Logistic Regression, Random Forests and Feed-forward Neural Networks (FNNs), each of which aims to predict the code from the clinical notes. In addition, an extensive number of experiments will be applied to different settings of the tested classification algorithm. Apart from using word embedding to transform a patient's free-text EHR into information that could be used to predict ICD-9 codes, this research will evaluate the impact of word embedding trained from MIMIC-III \cite{johnson2016mimic} dataset and medical domain word embedding. In short, this paper aims to provide a baseline for the learning-based ICD-9 code assignment on MIMIC-III dataset.

The contributions of this study are three-fold. The first contribution is the development of deep learning-based algorithms to map ICD-9 codes to clinical discharge summaries. The implementation of this research outperformed existing works under certain evaluation metrics. The second is the comparison of the performance of a wide variety of state-of-the-art machine learning and deep learning algorithms on MIMIC-III dataset. The third is the utilization of a set of standard metrics to assess the performance of ICD-9 code assignment on MIMIC-III dataset, which can be used as a baseline for further research.   

\section{Related Work} \label{sec:related-work}
The task of automatic ICD-9 coding has been attempted for decades. In 1995, Larkey and Croft \cite{larkey1995automatic} designed classifiers for the automatic assignment of ICD-9 codes to discharge summaries. Automated ICD-9 coding for radiology reports was one of the first challenges in informatics community \cite{pestian2007shared} in 2007. There are two major categories of approaches for automatically assigning ICD-9 codes using text-free clinical notes. One category is rule-based and the other category is learning-based. Rule-based systems are designed by human experts. This approach has outperformed other methods in many cases \cite{pestian2007shared, goldstein2007three}. However, this kind of system relies heavily on the manual intervention of medical professionals, thus making it difficult to maintain and scale up to more general cases. Learning-based systems, on the other hand, do not require any domain knowledge from medical experts and rely only on learning algorithms to find the underlying distribution of the provided datasets \cite{pakhomov2006automating, medori2010machine, ribeiro2001experimental}. A detailed review of extracting information from textual documents in the EHR can be found in \cite{meystre2008extracting} and \cite{ling2017methods}.  

End-to-End data-driven approaches have gained popularity in the last few years. Recent methods based on deep learning have also demonstrated state-of-the-art performance in a wide variety of tasks, including computer vision \cite{krizhevsky2012imagenet}, speech recognition \cite{graves2013speech}, and NLP \cite{collobert2011natural}. In the clinical domain, Choi el. al. \cite{choi2016using} used Recurrent Neural Networks (RNNs) to predict heart failures. Lipton el. al. \cite{lipton2015learning} utilized Long Short-Term Memory (LSTM) to classify 128 diagnoses from 13 frequently but irregularly sampled clinical measurements extracted from structured EHR data. Similarly, \textit{DoctorAI} \cite{choi2016doctor} and \textit{RETAIN} \cite{choi2016retain} utilized RNNs on structured EHR data for diagnostic classification. Many researchers also used deep learning on unstructured free-text to predict the diagnosis. Bai, for instance, proposed a deep transfer learning framework for ICD-9 coding by making use of a large number of MeSH domain knowledge\cite{zeng2019automatic}. Prakash et. al. \cite{prakash2017condensed} exploited raw text from Wikipedia as a source of knowledge and introduced condensed memory neural networks to learn the diagnosis on MIMIC-III data. Given that Prakash et.al. \cite{prakash2017condensed} tackled a problem similar to ours, our results were compared with their findings in Section \ref{sec:paper_comp}. A survey of recent deep learning techniques for EHR can also be found in \cite{shickel2018deep}.     

\section{Methodology}

\begin{figure}[ht]
\centering
\includegraphics[width=0.98\linewidth]{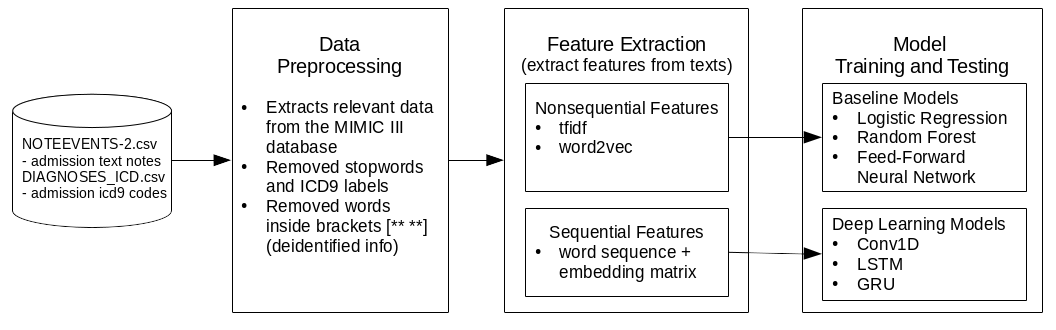}
\caption{Methodology Pipeline Overview}
\label{fig:pipeline}
\end{figure}

Figure \ref{fig:pipeline} depicts an overview of the methodology pipeline of this research. Our methodology involves the following steps: data preprocessing, feature extraction, and model training and testing. Specifically, the libraries used were: Spark for data preprocessing; Spark, Sklearn, and Gensim for feature extraction; and Spark ML, Keras for model training and testing. Azure virtual machines (NC24 with K80 GPU) were used to run our experiments. Sections \ref{sec_datapreproc} to \ref{sec_models} describe each step in more detail. Each model was evaluated under a set of metrics, as described in Section \ref{sec_metrics}. 

\subsection{Data Preprocessing}
\label{sec_datapreproc}
The MIMIC-III dataset is a large dataset relating to patients admitted to critical care units at a large tertiary care hospital. It contains de-identified medical records of patients who stayed from 2001 to 2012 within the intensive care units at Beth Israel Deaconess Medical Center \cite{johnson2016mimic}. The goal of this study is to explore useful semantic information using unstructured data. Therefore, only the free-text clinic note section from the dataset was used, specifically the \textit{noteevents} table. Furthermore, the focus was on the \textit{discharge summaries} category as it contained actual ground truth and free-text compared to other categories. Because \textit{discharge summaries} were written after carrying out the diagnosis, the notes were sanitized by removing any mention of class-labels (ICD-9 codes). This approach is similar to the one utilized by Prakash et al. \cite{prakash2017condensed}. 

Table \ref{tab:mimic-iii-stats} describes the number of unique patients, hospital admissions, ICD-9 codes and ICD-9 categories involved in MIMIC-III dataset. \textit{All MIMIC-III} describes the whole dataset, while \textit{noteevents} and \textit{discharge summaries} explain the corresponding subsets.

\begin{table}[htbp]
\begin{center}
\resizebox{\columnwidth}{!}{
    \begin{tabular}{| l*{4}{c} |}
        \hline
        Coverage & Patients & Hospital Admissions & ICD-9 Codes & ICD-9 Categories \\ \hline
        \textit{All MIMIC-III} & 46520 & 58976 & 6984 & 943 \\ \hline
        \textit{noteevents} & 46146 & 58361 & 6967 & 943 \\ \hline
        \textit{discharge summaries} & 41127 & 52726 & 6918 & 942 \\ \hline
    \end{tabular}
}
\caption{MIMIC-III Descriptive Statistics\label{tab:mimic-iii-stats}}
\end{center}
\end{table}

The data were preprocessed to produce separate datasets using two approaches. The first approach is to treat the ICD-9 code independently from each other, find the admissions (unique HADM\_ID) for each ICD-9 classification, and consider only records related to the top 10 and top 50 common ICD-9 codes. The top 10 and top 50 were chosen because they covered a majority of the dataset (76.9\% and 93.6\% as illustrated in Table \ref{tab:dataset-stats}). The second approach is to group ICD-9 codes into categories based on their hierarchical nature, with categories for larger sets of similar health conditions (For instance, "cholera due to vibrio cholerae" has the ICD-9 code 001.0, and is categorized as a type of cholera, which is also a type of intestinal infectious disease). The next step is to find the patients for top 10 and top 50 common categories. Evaluations would be separately performed on the four datasets, which will hereby be referred to as \textit{top-10-code}, \textit{top-50-code}, \textit{top-10-cat} and \textit{top-50-cat}, respectively.  

%
%

Table \ref{tab:top10-icd9} shows the top 10 ICD-9 codes and top 10 ICD-9 categories. Table \ref{tab:dataset-stats} also describes the number of unique hospital admissions related to the four datasets mentioned in the previous paragraph.

\begin{table}[htbp]
    \begin{center}
        \resizebox{\columnwidth}{!}{%
        \begin{tabular}{| l*{1}{c} |} \hline
            ICD-9 Code & Admissions \\ \hline 
            4019:  Hypertension & 20046 \\
            4280: Congestive heart failure & 12842 \\
            42731: Atrial fibrillation & 12589 \\
            41401: Coronary atherosclerosis & 12178 \\
            5849: Acute kidney failure & 8906 \\
            25000: Diabetes Type II & 8783 \\
            2724: Hyperlipidemia & 8503 \\
            51881: Acute respiratory failure & 7249 \\
            5990: Urinary tract infection & 6442 \\
            53081: Esophageal reflux & 6154 \\
            \hline
        \end{tabular}
        \quad
        \begin{tabular}{| l*{1}{c} |} \hline
            ICD-9 Category & Admissions \\ \hline
            401: Essential hypertension & 20646 \\
            427: Cardiac dysrhythmias & 16774 \\
            276: Disorders of fluid electrolyte  & 14712 \\
            272: Disorders of lipoid metabolism & 14212 \\
            414: Other chronic ischemic heart disease & 14081 \\
            250: Diabetes mellitus & 13818 \\
            428: Heart failure & 13330 \\
            518: Other diseases of lung & 12997 \\
            285: Other and unspecified anemias & 12404 \\
            584: Acute kidney failure & 11147 \\
            \hline
        \end{tabular}
        }
    \caption{Admission number for Top 10 ICD-9 codes and top 10 ICD-9 categories\label{tab:top10-icd9}}
    \end{center}
\end{table}

\begin{table}[htbp]
    \begin{center}
        \resizebox{0.8\columnwidth}{!}{
        \begin{tabular}{| l*{2}{c}|} \hline
            Data Set & Hospital Admissions & \textit{discharge summaries} Coverage (\%) \\ \hline 
            \textit{top-10-code} & 40562 & 76.93\% \\ \hline
            \textit{top-50-code} & 49354 & 93.60\% \\ \hline
            \textit{top-10-cat} & 44419 & 84.24\% \\ \hline
            \textit{top-50-cat} & 51034 & 96.79\% \\ \hline
            \end{tabular}
        }
            \caption{Dataset Descriptive Statistics\label{tab:dataset-stats}}
    \end{center}
\end{table}

The filtered datasets will be split into 50-25-25 for training, validation and testing.

\subsection{Feature Extraction}
\label{sec_fe}
Two approaches will be used for feature extraction: They include Term Frequency - Inverse Document Frequency (\textit{tfidf}) and \textit{word2vec} \cite{mikolov2013distributed}. The \textit{tfidf} serves as a baseline of comparison with \textit{word2vec}.

The \textit{tfidf} aims to evaluate the level of importance of a word to a document in a collection of documents or corpus. It is the product of two statistics: \textit{tf} and \textit{idf}. While \textit{tf} is the number of times a word appears in a given document, and \textit{idf} measures whether a word is common or rare across the corpus. The following definition of \textit{idf} will be used for our calculations:

\begin{equation*}
idf(w) = log\frac{n_{d}}{df(d,w)} + 1
\end{equation*}

where $n_d$ is the total number of documents, and $df(d,w)$ represents the number of documents that contain the word $w$.

To calculate \textit{tfidf}, all the notes in the filtered training data set were first tokenized. Next, a document-word matrix with the count of each word in each note (\textit{tf}) was created. Finally, each word was multiplied by the corresponding \textit{idf}. Two \textit{tfidf} configurations were also used: (1) one with top 40,000 words with highest \textit{tfidf} scores as the bag of word features; (2) the other one with a minimum document frequency of 10 and a maximum document frequency of 0.8 of the total number of documents, which reduced the total number of words to around 20,000 words.

The model \textit{word2vec} takes a tokenized text corpus as an input and produces word vectors as an output. The Continuous Bag of Words (CBOW) architecture was then used to predict the target word based on the context: words that precede and follow the target word. The CBOW is basically a Neural Network model that consists of inputs, projection and output layers where the traditional non-linear hidden layer is removed to reduce the time complexity and the projection layer is shared by all the words. The inputs are words in the context. We used text notes from MIMIC-III as corpus to train our \textit{word2vec} model. Pre-trained word vectors induced from PubMed were also utilized. PubMed is a database of biomedical literature and the word vectors can be found at \href{https://github.com/cambridgeltl/BioNLP-2016}{https://github.com/cambridgeltl/BioNLP-2016} \cite{Chiu2016}

\subsection{Model Training and Testing}
\label{sec_models} 
One fundamental assumption adopted by traditional supervised learning algorithms is that each sample has only one label assigned to it. In our problem, each sample has multiple (one or more) ICD-9 codes attached to it. Generally, there are two main methods for tackling the multi-label classification problem \cite{tsoumakas2007multi} One is the problem transformation methods and the other is algorithm adaptation methods. Problem transformation methods transform the multi-label problem into a set of binary classification or regression problems, and multiple binary classifiers are trained separately for each label. Algorithm adaptation methods, on the other hand, adapt the algorithms to perform multi-label classifications in its full form, and only one classifier is trained for all the labels.

In our study, three baseline approaches were first created: Linear Regression, Random Forests and Feed-forward Neural Networks. Then problem transformation methods were used to obtain the multi-label output for Linear Regression and Random Forests classifiers. Specifically, in order to assign each sample a set of target labels, $n$ different models for $n$ different labels were trained. Each model independently predicts a mutual exclusive output (0 or 1) for each sample data. For Feed-Forward Neural Networks, algorithm adaptation based methods were utilized, given that the neural network could be easily adapted to multi-label problem by setting up multiple neurons in the network output layer and each neuron represents a target label correspondingly. Similar to FNNs, algorithm adaptation-based methods were used in our deep learning models. In the following sub-sections, our implemented models will be described in detail. 

\subsubsection{Baseline Models}
\textbf{Logistic Regression (LR):} Our first baseline model is a binomial logistic regression model implemented using Spark ML. For each label (ICD-9 code or category), a separate logistic regression model was trained, and each model independently predicted the said label (0 or 1 for the corresponding ICD-9 code or category). Different configurations were tried; specifically, ``the number of iterations" was tuned between 5 to 100. Because only notes under \textit{discharge summaries} category were used, there was one note per admission. Features extracted from this note were used as inputs for this classifier. For \textit{tfidf}, the features were directly used as input features. For \textit{word2vec}, the input features were the average of all the feature vectors of the words in the notes. This simple yet popular method was successfully applied in other studies to obtain sentence or document embedding \cite{faruqui2015retrofitting, kenter2015ad}.

\textbf{Random Forests (RFs):} Our second baseline model is a random forest model implemented using Spark ML. The same approach and input for the logistic regression were used here (one model for each label). Different configurations were also evaluated; specifically the ``tree depth" was tuned between 5 to 30.

\textbf{Feed-forward Neural Networks (FNNs):} One advantage of Neural Networks is that it can be fitted to multi-label problems in just one model with the proper activation function. The FNNs were implemented as the baseline for algorithm adaptation based multi-label classification problem (see Section \ref{sec_models}). The same input features and train-test data split as previously described were used. The ReLU activation function was utilized for all the hidden layers and sigmoid function was used for the output layer, binary cross entropy as the loss function, and stochastic gradient descent as the optimizer. Several neural network models with one to four different hidden layers were also tried. For each hidden layer, a total of seven models was employed with the following combination of neuron sizes: 50, 100, 300, 500 and 1000. Among our model architectures, the best performed model pipeline is shown in Table \ref{tab:ff-nn}.

\begin{table} [htbp]
    \begin{center}
        \resizebox{0.7\columnwidth}{!}{%
        \begin{tabular}{| l*{4}{c}|}        
            \hline
            Layer  & 0  & 1 & 2 & 3 \\
            \hline
            NN     & input & dense (ReLU) & dense (ReLU) & dense (ReLU) \\ 
            Para    & - & 5000 & 500 & 100 \\
            \hline
            
        \end{tabular}
        }
        \caption {Configuration Details of the Best Performed FNN Architecture.} \label{tab:ff-nn}

    \end{center}
\end{table}

\subsubsection{Deep Neural Network Models}
In this study, the problem of ICD-9 code assignment from clinical notes was treated as multi-label classification problem on sequential observations $x_1, x_2, \cdots, x_n$, where $x_i$ is the \textit{word2vec} features calculated for word $i$ in the discharge summary. Unlike the features used for the baseline models in which the sequential information was not preserved, each word was taken sequentially from the discharge summary. The input features for this classifier are \(N\) most recent word sequences taken from the notes. If there were insufficient feature events, zero vectors were padded at the beginning. The word sequence was then converted into vectors using an embedding matrix based on a \textit{word2vec} model (See Section \ref{sec_fe}). 

\begin{figure}[htbp]
\centering
\includegraphics[width=0.75\linewidth]{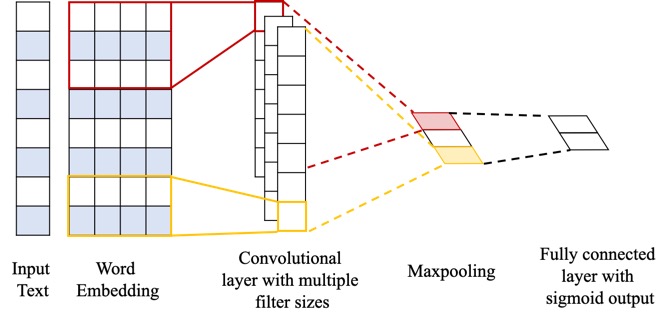}
\caption{CNN Architecture}
\label{fig:conv1d}
\end{figure}

\textbf{Convolutional Neural Networks} (CNNs) have achieved remarkable results in image processing related problems. In recent years, CNN models have shown excellent results for NLP such as in semantic parsing \cite{yih2011learning}, search query retrieval
\cite{shen2014learning}, and sentence classification \cite{kim2014convolutional}. Thus, a series of experiments with CNNs were carried out. In general, the same architecture described in \cite{kim2014convolutional} was applied. As shown in Figure \ref{fig:conv1d}, the features were first concatenated into $n \times k$ feature vector, where $n$ is the number of words, and $k$ is the number of dimensions extracted from \textit{word2vec}. A set of convolution filters with dimension $h \times k$ was then applied to a window of $h$ words to produce new features. The filters were then applied to each possible window of words in the sentences to produce a feature map. Finally, a max-overtime pooling operation over the feature map was applied to generate the fully connected layer. A sigmoid activation function was also applied to generate the multi-label output.    

Various number of layers for our CNNs were tried, including three to ten convolutional (conv) layers with size 64, 128, or 256 for each layer. Then, each layer was followed with a max pooling layer, and one to three fully connected (fc) dense layers were attached to the last convolutional layer with a size of 4096, 1024 or 128. Among our model architecture setting, the best performed model pipeline is shown in Table \ref{tab:conv1d}. The same architecture was used on both \textit{top-10} and \textit{top-50} codes. Based on the hardware setting described, the training time for CNNs was less than 30 minutes with 500 maximum epochs and early stop if the validation loss did not improve for consecutive 10 epochs. 

\begin{table} [htbp]
    \begin{center}
        \resizebox{\columnwidth}{!}{%
        \begin{tabular}{| l*{8}{c}| r}        
            \hline
            Layer  & 0  & 1 & 2 & 3 & 4 & 5 & 6  & 7\\
            \hline
            Architecture     & input & conv & max pooling & conv & max pooling & conv & max pooling & fc \\ 
            Para    & - & 128 - 5 & 5 & 128 - 5 & 5 & 128 - 5 & 35 & 128 \\
            \hline
            
        \end{tabular}
        }
        \caption {Configuration Details of the Best Performed CNN Architecture.} \label{tab:conv1d}

    \end{center}
\end{table}

\begin{figure}[htbp]
\centering
\includegraphics[width=0.75\linewidth]{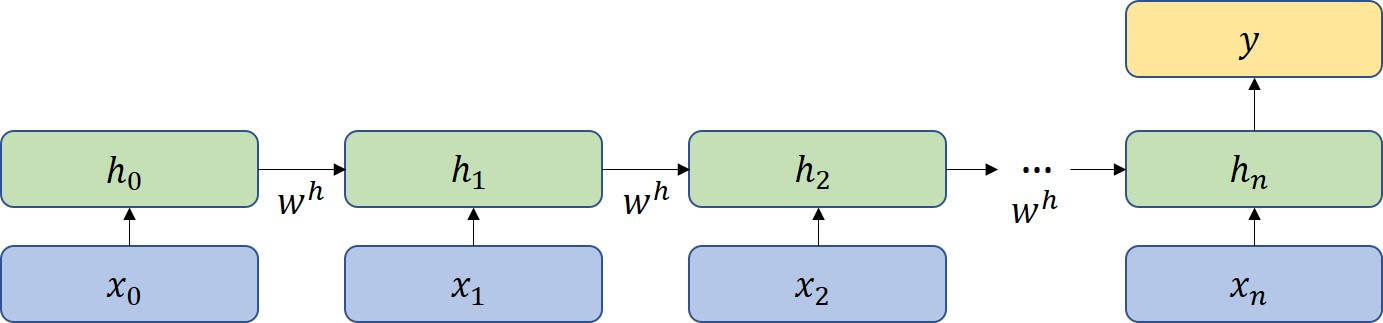}
\caption{RNNs architecture}
\label{fig:rnn-abstract}
\end{figure}

\textbf{Recurrent Neural Networks (RNNs):} RNNs are a type of neural network architecture designed to handle sequential inputs. They have shown promising results in many machine learning tasks \cite{graves2012supervised}. Several RNN architectures were explored in this study. All the architectures  follow the same pattern shown in Figure \ref{fig:rnn-abstract}, where blue circles represent the text feature vectors. The green rectangles and the yellow rectangle represent recurrent hidden layers and the multi-label code assignment, respectively. Basically, the RNN cells went through the input sentences. Each word $x_t$ in the sentences generated a hidden layer $h_t$. Each hidden layer was connected with a directed connection weights $w_h$ to its successive layer $h_{t+1}$. The weights $w^h$ were shared among all the hidden layers (shared over time). The hidden layers of the RNN generated the outputs $\hat{y}$ when the RNN cells reached the last word. Sigmoid cross-entropy was then used as the loss function and RMSprop as the optimizer.

\begin{equation*}
loss(\hat{y}, y) = -\frac{1}{N}\sum_{n=1}^{l=N}y_n \cdot log(\hat{y_n}) + (1-y_n) \cdot log(1-\hat{y_n})
\end{equation*}

Although RNNs can handling input sequences of variable sizes, in practice, they face difficulties when modeling long-term dependencies \cite{bengio1994learning}. To address this issue, various recurrent units were developed. Among those sophisticated recurrent units, in this study, two popular ones were evaluated: LSTMs \cite{hochreiter1997long} and Gated Recurrent Units (GRUs) \cite{cho2014properties}. Both of them could capture sequence-based inputs with long-term dependencies by utilizing a memory mechanism. Generally, LSTM contains three gates: the input gate, forget gate and output gate. The forget gate decides what information from previous and current input should be preserved or ignored. The input gate decides the values that will be updated to the cell states. The output gate decides what the next hidden state should be. A GRU is an LSTM without an output gate, and it uses an update gate to decide what past information should be kept and a reset gate decides how much past information should be discarded.

The input features to RNNs are the same as those used in CNNs. Three stacked recurrent layers with a combination 64, 128 and 256 units for each layer in our RNNs were tried. To predict the ICD-9 classification, only the output nodes of the last time step were considered, and the same activation function and loss function were applied as with the NNs. The best performed mode architecture for LSTMs and GRUs are shown in Table \ref{tab:rnn}. Both architectures that performed best have two stacked recurrent layers with the same unit numbers for each layer. Based on the specified hardware setting, the training time was about 6 hours for GRUs and 18 hours for LSTMs with 200 maximum epochs and early stop if the validation loss did not improve for consecutive 5 epochs. 

\begin{table} [htbp]
    \begin{center}
        \resizebox{\columnwidth}{!}{%
        \begin{tabular}{| l*{5}{c}|| l*{5}{c} | r}        
            \hline
            Layer  & 0  & 1 & 2 & 3 & 4 & Layer & 0 & 1  & 2 & 3 & 4 \\
            \hline
            LSTMs  & input  & lstm & dropout & lstm & dropout & GRUs & input & gru & dropout & gru & dropout\\ 
            Para    & - & 256 & 0.5 & 64 & 0.5 & Para & - & 256 & 0.5 & 64 & 0.5 \\
            \hline
            
        \end{tabular}
        }
        \caption {Configuration Details of the Best Performed LSTMs and GRUs Architectures.} \label{tab:rnn}

    \end{center}
\end{table}

\subsection{Metrics}
\label{sec_metrics}
The combinations of our dataset, feature extraction methods, and models are evaluated under different performance metrics, including precision, accuracy, F-score and recall metrics for multi-label classification. Specifically, the following metrics are used \cite{zhang2014review}:
\begin{align*}
 \text{Precision}  &= \frac{1}{n}\sum^n_{i=1}\frac{\lvert Y_i \cap Z_i \rvert}{\lvert Z_i \rvert}   &   \text{Recall}  &= \frac{1}{n}\sum^n_{i=1}\frac{\lvert Y_i \cap Z_i \rvert}{\lvert Y_i \rvert} \\
 \textit{$F_1$} &= \frac{1}{n}\sum^n_{i=1}\frac{2 \lvert Y_i \cap Z_i \rvert}{\lvert Y_i \rvert + \lvert Z_i \rvert}       &   \text{Accuracy} &= \frac{1}{n}\sum^n_{i=1}\frac{\lvert Y_i \cap Z_i \rvert}{\lvert Y_i \cup Z_i \rvert} 
\end{align*}

where $Y_i$ is the set of predicted labels, $Z_i$ is the set of ground truth labels, and $n$ is the number of samples. Basically, precision calculates the proportion of predicted labels that are correct. Recall calculates the proportion of the actual labels that are correctly predicted. $F_1$ is the harmonic mean of precision and recall. Accuracy is the average proportion of the predicted correct labels to the total number of labels for all instances.

\section{Results}
This section illustrates the performance in three different aspects: (1) the baseline results, (2) the performance under different configurations, and (3) the best model performance.

\subsection{Model Performance under Different Configurations }
\label{sec_modelperf_diffconfig}
Different model configurations have been tried to give us insight into the most appropriate model configuration. Table \ref{tab:fe-methods} describes the different methods of feature extraction used and the parameters tweaked. The features extracted are divided into two categories: non-sequential and sequential features. The non-sequential features include \textit{tfidf} and \textit{word2vec}, both of which were used in Logistic Regression, Random Forests, and NNs. The sequential features includes \textit{word2seq} (word sequences) used in conjunction with an embedding matrix based on \textit{word2vec}, which were used in CNNs, RNNs, LSTMs, and GRUs. It is pertinent to note that we experimented on (1) using our custom \textit{word2vec} model created from the MIMIC-III dataset and (2) utilizing pre-trained word vectors obtained from PubMed \cite{Chiu2016}. The vectors for stop words in the embedding matrix are all zeros.

\begin{table}[ht]
\centering
\resizebox{\columnwidth}{!}{%
\begin{tabular}{| c |l p{20em}|}
\hline
Feature Extraction & Configuration & Value \\ \hline
\multirow{3}{*}{tfidf}  & feature size & 20301 - 40000  \\ \cline{2-3}
 & minDocFreq & 3 - 10  \\ \cline{2-3}
  & max\_df &  0.8 - 1.0 \\ \hline
\multirow{3}{*}{word2vec} & database & self trained from MIMIC-III (m3) or pre-trained from Pubmed (pm) \cite{Chiu2016} \\ \cline{2-3}
 & feature size & 100 - 600  \\ \cline{2-3}
 & pre-trained config &  context window size 2 (win 2) or 30 (win30) \cite{Chiu2016}\\ \hline
\multirow{3}{*}{wordseq} & sequence length & 1500-2000 \\ \cline{2-3}
& stopwords & removed from sequence or not removed  \\ \cline{2-3} 
& embedding matrix & derived from the word2vec under different configurations \\ \hline
\end{tabular}
}
\caption{Feature Extraction Methods \label{tab:fe-methods} }
\end{table}

\begin{figure}[htbp]
\centering
\subfigure[Nonsequential]{\includegraphics[width=0.47\linewidth]{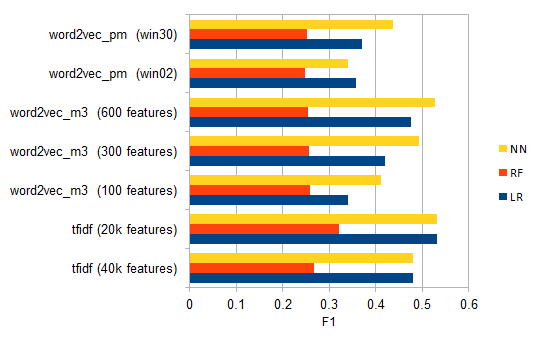}\label{fig:model-perf-diff-cfg-nonseq}}
\subfigure[Sequential]{\includegraphics[width=0.47\linewidth]{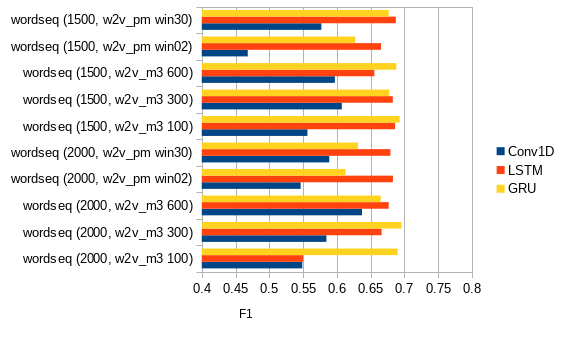}\label{fig:model-perf-diff-cfg-seq}}
\caption{Model Performance under Different Configurations}
\label{fig:model-perf-diff-cfg}
\end{figure}

Figure \ref{fig:model-perf-diff-cfg} indicates the model performance of each model using different feature extraction methods on the \textit{top-10-code} dataset. For each model, the configuration that provided the best performance here is used on the \textit{top-50-code}, \textit{top-10-cat}, and \textit{top-50-cat} datasets. The results are further explained in the next section (see Section \ref{sec_bestmodelperf}).

Our three non-sequential models (Logistic Regression, Random Forests and NNs) had different but comparable performance depending on the features used. For example, for \textit{top-10} code classification, \textit{tfidf with 20k features} produced the best $F_1$ results of 0.532 and 0.322 for Logistic Regression and Random Forests respectively. However,  \textit{word2vec\_m3 with 600 features} produced the best results for FNNs with $F_1$ of 0.528 (although \textit{tfidf} also gave a fairly good result for FNNs with $F_1$ of 0.488). Thus, \textit{tfidf} configurations generated better results than those of \textit{word2vec}. It is possible that the \textit{word2vec} features lost information as the average of the word vectors was used to obtain document embedding for the non-sequential models. However, note that NNs gave fairly good results ($F_1$ 0.528) using the \textit{word2vec} feature which was at most 600 dimensions compared to \textit{tfidf} which was 20k or above. With a larger feature size, it is reasonable to say that \textit{tfidf} kept a better global representation of document embedding. This ability could be used to explain the code classification. However, \textit{word2vec} did retain the satisfactory representation of word embedding despite the substantial reduction in feature dimension (i.e., 20k down to 600).

Four sequential models (CNNs, simple RNNs, LTSMs and GRUs) were run under different configuration with two different types of embedding matrices (our self-trained \textit{word2vec} from MIMIC-III corpus and pre-trained \textit{word2vec} from PubMed \cite{Chiu2016}) and two sentence sequence lengths (1500 and 2000). In the \textit{top-10} code classification, \textit{seq. length 2000 + word2vec\_m3 w/ 600 features} generated the best $F_1$ result for CNNs, \textit{seq. length 1500 + word2vec\_pm (win30)} for LSTMs, and \textit{seq. length 2000 + word2vec\_m3 w/ 300 features} for GRUs. In short, all feature extraction methods generated good and comparable results for CNNs, LSTMs, and GRUs. Our self-trained \textit{word2vec} also performed fairly well compared with the pre-trained \textit{word2vec} models from PubMed. The best performed models with self-trained \textit{word2vec} have $F_1$ of 0.637 (CNNs), 0.696 (GRUs) and 0.683 (LSTMs) while the best performed models from PubMed \textit{word2vec} are 0.589 (CNNs), 0.677 (GRUs), and 0.687 (LSTMs). Under different configurations, our self-trained $word2vec$ outperformed PubMed \textit{word2vec} in most cases.    
Experiments were also performed for simple RNNs. The results, however, were poor (0.0 - 0.08 $F_1$ at best). More information on this is explained in detail in Section \ref{sec:discussion}. In addition, our top $F_1$ scores are linked to GRUs and LSTMs with GRUs providing slightly better results.

\subsection{Best Model Performance}
\label{sec_bestmodelperf}

\subsubsection{Overview}
\begin{figure}[ht]
\centering
\subfigure[top-10-code]{\includegraphics[width=0.47\linewidth]{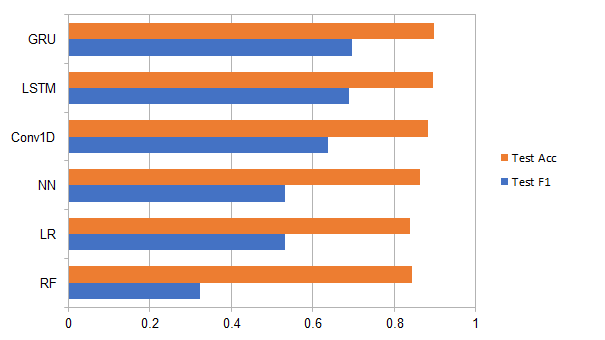}\label{fig:model-perf-top-10-code}}
\subfigure[top-10-cat]{\includegraphics[width=0.47\linewidth]{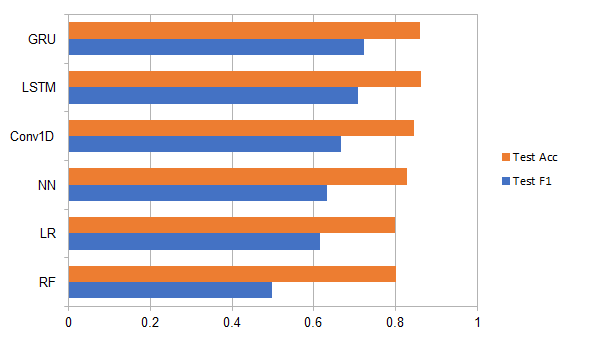}\label{fig:model-perf-top-10-cat}}
\caption{Model Performance Top 10}
\label{fig:model-perf-top10}
\end{figure}

\begin{figure}[ht]
\centering
\subfigure[top-50-code]{\includegraphics[width=0.47\linewidth]{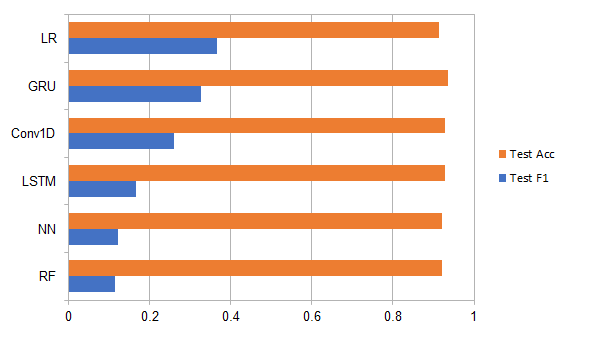}\label{fig:model-perf-top-50-code}}
\subfigure[top-50-cat]{\includegraphics[width=0.47\linewidth]{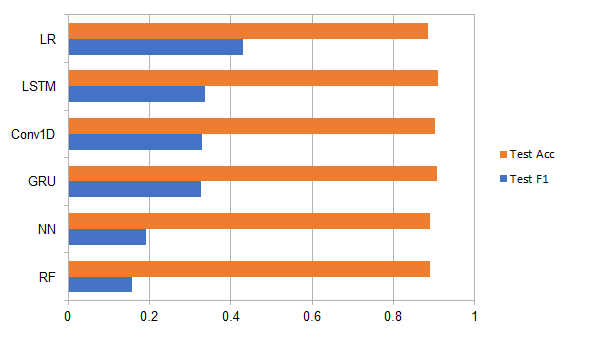}\label{fig:model-perf-top-50-cat}}
\caption{Model Performance Top 50}
\label{fig:model-perf-top50}
\end{figure}

\begin{figure}[ht]
\centering
\subfigure[top-50-code]{\includegraphics[width=0.47\linewidth]{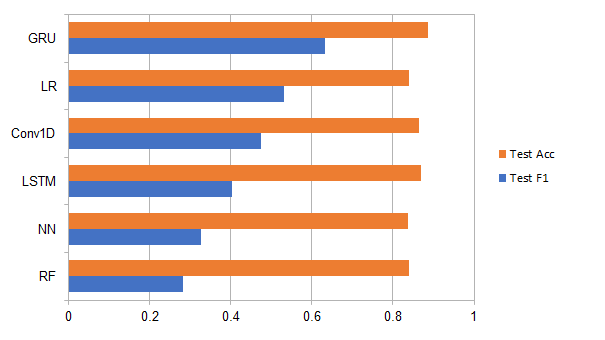}\label{fig:model-perf-top-50-code-first10}}
\subfigure[top-50-cat]{\includegraphics[width=0.47\linewidth]{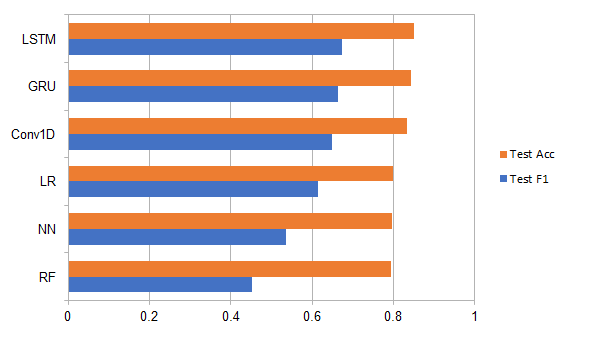}\label{fig:model-perf-top-50-cat-first10}}
\caption{Model Performance Top 50 (first 10 labels only)}
\label{fig:model-perf-top50-first10}
\end{figure}

Figures \ref{fig:model-perf-top10} and \ref{fig:model-perf-top50} show the model performance, which is ordered from the best to worse, for the \textit{top-10-code}, \textit{top-10-cat}, \textit{top-50-code}, and \textit{top-50-cat} dataset. Figure \ref{fig:model-perf-top50-first10} indicates the model performance for \textit{top-50-code} and \textit{top-50-cat} considering only the first 10 labels. Raw data are also shown in Tables \ref{tab:model-perf-top-10-code} to \ref{tab:model-perf-top-50-cat-first10} in the Appendix.


For \textit{top-10-code} and \textit{top-10-cat}, GRUs generated the best $F_1$ results (of 0.6957 and 0.7233, respectively). Hence, \textit{top-10-cat} generated slightly better results than did \textit{top-10-code}. This makes sense because there are more samples per label in \textit{top-10-cat} and because the labels are less specific (the differences between labels are larger). Logistic Regression and Random Forests overfit the data with about 80\% to 95\% training $F_1$ but about 40\% to 50\% less on testing data). Even though FNNs are not overfitting, the results on testing data are slightly better than other baseline models. CNNs produces better results than FNNs, but there are more significant improvements with LSTMs and GRUs (about 70\% $F_1$). This result signifies that our LSTMs and GRUs model can extract information from the sequence of words, thereby improving the $F_1$ and the result accuracy.

For \textit{top-50-code}, Logistic Regression generated the best $F_1$ result of 0.3662. However, if only the first 10 labels are considered, GRUs generated the best $F_1$ result at 0.6328. For both all the label results and the first 10 label results, GRUs generated the best precision and accuracy results of 0.7520 and 0.8871, respectively.
For \textit{top-50-cat}, Logistic Regression also generated the best $F_1$ result of 0.4301. However, if only the first 10 labels are considered, LSTMs generated the best $F_1$ result of 0.6738. For both all the label results and the first 10 label results, GRUs generated the best precision and accuracy results of 0.7515 and 0.8345, respectively. Hence, \textit{top-50-cat} generated slightly better results than did \textit{top-50-code}. The baseline models (Logistic Regression and Random Forests) also overfit here.

\subsubsection{Precision-Recall Curve}
\begin{figure}[htbp]
\centering
\subfigure[top-10-code]{\includegraphics[width=0.49\linewidth]{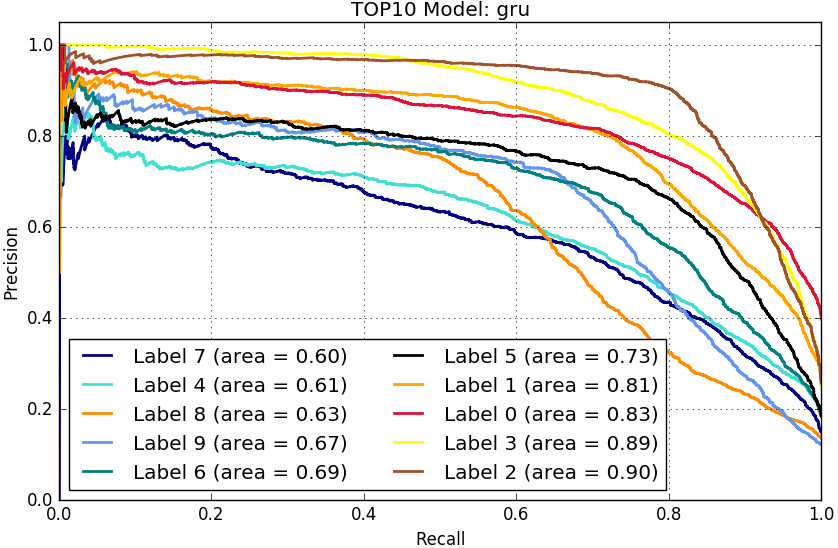}\label{fig:curve_gru_10}}
\subfigure[top-50-code]{\includegraphics[width=0.49\linewidth]{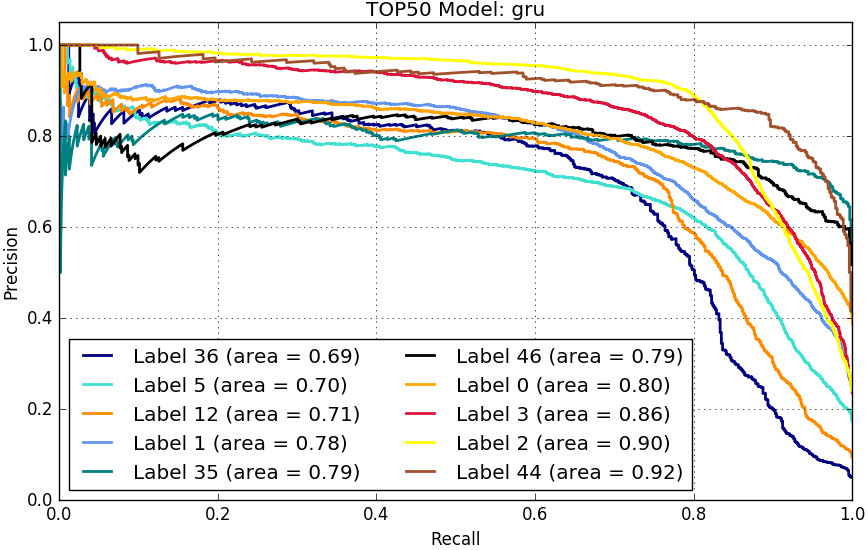}\label{fig:curve_gru_50}}
\caption{Class-wise Precision-recall Curve for the Top 10 and the Best-performed 10 Labels for Top 50 Code Classification.}
\label{fig:curve}
\end{figure}

Table \ref{tab:average_precision_performance} presents the average overall precision performance of our selected best performance models for GRUs, LSTMs and CNNs. Average Precision (AP) summarizes the precision-recall curve as the mean of precisions achieved at different recall values and it is calculated as follows:

\begin{equation*}
    AP = \sum_n (R_n - R_{n-1})P_n
\end{equation*}

where $P_n$ and $R_n$ are the precision and recall at the nth threshold. As shown in the table below, GRUs generated the best precision results for \textit{top-10-code} and \textit{top-50-code}. Figure \ref{fig:curve} shows precision-recall curve for the best-performed models for each label in \textit{top-10-code} and the best-performing 10 labels for \textit{top-50-code}. In this picture, it can be seen that for \textit{top-10-code}, the top five common codes performed better than did the later ones. For example, the third common label (label 2: 42731), atrial fibrillation, has the highest AP of 0.90. followed by coronary atherosclerosis (label 3: 41401) with AP of 0.89, hypertension with AP of 0.83 (label 0: 4019) and congestive heart failure (label 1: 4280) with AP of 0.81. However, in the performance of \textit{top-50-code}, it was observed that less common codes could also achieve high AP scores. For example, label 44: 7742, neonatal jaundice associated with preterm delivery, which have 2183 samples in the training dataset has the highest AP score of 0.92, and label 46: V053 with 2119 samples reached fifth. Other class-wise precision-recall curve for our tested models can be found in the Appendix.

\begin{table}[ht]
\centering
\resizebox{\columnwidth}{!}{%
\begin{tabular}{|l*{6}{c} |}
\hline
Model & \textit{top-10-code} & \textit{top-10-cat} & \textit{top-50-code} & \textit{top-50-cat} & \textit{top-50-code(first10)} & \textit{top-50-cat(first10)} \\ \hline
\textit{LSTMs} 	&0.7243&0.7915& 0.3715 &0.4929 &0.7571&0.8426\\ \hline
\textit{GRUs} 	&0.7362&0.7849& 0.4518 &0.4792 & 0.7949&0.8425\\ \hline
\textit{CNNs} &0.6719&0.7293& 0.3757 & 0.4565& 0.7424&0.8269\\ \hline
\end{tabular}
}
\caption{Average Precision Performance}
\label{tab:average_precision_performance}
\end{table}

\subsubsection{Results Comparison}
\label{sec:paper_comp}
Prakash and Zhao \cite{prakash2017condensed} used bag-of-words from discharge notes and Condensed Memory Neural Networks (C-MemNN) to tackle the same problem in this research. They tested their algorithm with top 50 and top 100 labels under metrics such as the macro average of Area Under the Curve (AUC), average precision over the top five predictions (Precision @5), and hamming loss. 

\begin{equation*}
AUC_{macro} = \frac{1}{q}\sum^q_{j=1}AUC_j
\end{equation*}

\begin{equation*}
\textit{Hamming Loss} = \frac{1}{nq}\sum^n_{i=1}\sum^q_{j=1}xor(Z_{i,j},  Y_{i, j})
\end{equation*}

where $AUC_j$ is the $AUC$ for each label, $q$ is the number of labels, and $n$ is the number of samples. Macro $AUC$ is used to calculate the unweighted mean of the $AUC$ values for each label. Hamming loss represents the fraction of labels that are incorrectly predicted. To compare our work with theirs, the same metrics were used for our best performed models for top 50 codes (top 100 labels are not compared), and the results are presented in Table \ref{tab_memory_network_compare}. 

Based on the results, it can be seen that while their hamming loss is better than ours, our work outperforms theirs in terms of macro AUC on GRU models and significantly performs better on top five precision for all of our models.

\begin{table}[ht]
\centering
\resizebox{0.65\columnwidth}{!}{%
\begin{tabular}{|l*{3}{c}|}
\hline
Model & AUC (macro) & Precision @5 & Hamming Loss \\ \hline
\textit{C-MemNN \cite{prakash2017condensed}} & 0.833 &0.42 & \textbf{0.01} \\ \hline
\textit{GRUs} & \textbf{0.8599} &\textbf{0.8109} & 0.0645 \\ \hline
\textit{LSTMs} & 0.8298 & \textbf{0.8054} & 0.0714\\ \hline
\textit{CNNs} & 0.8302 & \textbf{0.7998} & 0.0714\\ \hline
\end{tabular}
}
\caption{Performance Comparison with Reference \cite{prakash2017condensed}\label{tab_memory_network_compare}}
\end{table}



\section{Discussion}
\label{sec:discussion}
Although some studies have used deep learning for automatic ICD-9 code assignment, a few of them have focused on getting meaningful information directly from unstructured clinical notes. Existing research on such tasks are difficult to compare with each other because there is a lack of standard comparison metrics. This study is significant in that it conducted extensive experiments to examine the performance of popular machine learning and deep learning algorithms under a set of standard metrics, precision, recall, $F_1$ and accuracy. As the output label vector for our task was sparse (for each sample, only a few labels were generally active), the influence of sparsity was considered when choosing the measurements. For instance, both hamming loss and accuracy are in favor of sparse labels. An all-zero classifier would have the hamming loss close to 0 and the accuracy close to 1. Precision, recall, and $F_1$ are more reasonable choices compared to hamming loss and accuracy. If the developed system is aiming for recommendation, $Precision @k$ has often been used for such purpose \cite{agrawal2013multi}. Other optimized measures for sparse labels are also worth exploring in future research \cite{dembczynski2012label}.

Our findings support the use of RNNs for code assignment on patient notes, as they show better performance than other machine learning models as well as previously published systems \cite{prakash2017condensed} under certain metrics. Except from the sequential RNN models, the other models only consider the current input and they have no notion of order in time. They simply can not remember anything about what happened in the past. In a FNN, the information moves straight through the network in one direction. Because of that, the information hardly touches a node twice. In contrast, RNNs try to derive relations from the current word and what it has learned from previous words in the same sentence. In RNNs, each node at a time step takes an input from the previous node using a feedback loop.

To further understand what form of deep neural network architecture works better on automatic ICD-9 coding, the performance of CNNs, simple RNNs, and RNNs with long-term dependencies capability were also compared. Among all the compared methods (including the baseline models), simple RNNs without memory mechanism produced the worst performance. The $F_1$ is 0.08 at best for various configurations. Although it is well known that RNNs can only use information from near past, the unsatisfactory results may be due to the learning ability of the network itself. As explained in \cite{bengio1994learning}, RNNs for tasks with long-term dependencies would not be robust to input noise or would not be efficiently trainable by gradient descent, and the gradient of the loss function decays exponentially with time. Worth noting is that in order to know if representations from future time steps would help the network to gain extra information, bidirectional LSTMs and GRUs were also implemented in our experiments. Bidirectional RNNs connects two hidden layers of opposite directions to the same output. Such mechanisms make the network receive information from past and future states simultaneously. Bidirectional RNNs are especially useful when the later context of the input is needed. It was observed that bidirectional RNNs do not outperform (comparable or slightly worse) the RNNs for the \textit{top-10 code} assignment task. Considering the extra computation cost of the bidirectional RNNs, this network architecture was not further explored with the belief that peeking future information may not beneficial to our task.

Despite the fact that CNNs did not perform as good as LSTMs and GRUs, the training time for CNNs (30 minutes) was significantly less than that of GRUs (6 hours) and LSTMs (18 hours). The computation for CNNs can happen in parallel, while RNNs need to be processed sequentially, After all, the subsequent steps depend on previous ones. As part of future work, the results may be improved with newly designed network architectures, such as Temporal Convolutional Networks (TCNs) \cite{lea2017temporal}. Bai et al. compared a series of benchmark competitions of TCNs versus RNNs, LSTMs, and GRUs across eleven different industry standard RNNs problems. They found that TCN models substantially outperformed generic recurrent architectures such as LSTMs and GRUs. They also showed that TCNs exhibited longer memory than did recurrent architectures with the same capacity \cite{BaiTCN2018}.

The paper also evaluates the impact of word embedding on the performance of our tested methods. For baseline models, a simple yet popular method to generate the non-sequential \textit{word2vec} feature vector was used. Specifically, the word vectors from a discharge summary were averaged. Models using the averaged word vector features performed worse than those using \textit{tfidf} features. While for deep learning, because the sequential information can be retained, almost all the networks with \textit{word2vec} could generate better results than our baseline models. In addition, compared to our self-trained \textit{word2vec} from MIMIC-III and \textit{word2vec} downloaded from PubMed, they showed comparable performance and the \textit{word2vec} trained from MIMIC-III generally produced slightly better results.

Although LSTMs and GRUs could capture long-term dependencies, the length of our input sequence could still be too long for LSTMs and GRUs to retain useful information. A different representation may be used to shorten the sequence, e.g. sentence2vec or paragraph2vec \cite{le2014distributed}. In addition to that, based on the comparison in Section \ref{sec:paper_comp}, memory networks provide better results for certain metrics. word2vec representation and memory network can help address this problem. 

Our current models for top 50 ICD-9 codes and categories were not successful. This failure may be because our current model design could not effectively distinguish between 50 different labels. To improve our model capability, we could run five \textit{top-10} models in parallel (each model predicting 10 labels), thereby making our \textit{top-50} models have the same model capability as our \textit{top-10} models. It was also observed that the samples for labels 11 to 50 were greatly imbalanced (i.e., the positive samples where very few compared to negative samples). Hence, the data might not be sufficient for the deep neural network to learn adequate useful representations. As part of future work, the sample size of the clinical notes that contains those labels needs to be balanced or increased. Zeng et al. \cite{zeng2019automatic} also showed that deep transfer learning could improve the performance of automatic ICD-9 coding on labels with limited samples.

Our custom \textit{word2vec} model used CBOW. Skip-gram (even though the pre-trained \textit{word2vec} induced from PubMed are skip-gram based) was not used. Some previous studies have noted that skip-gram outperforms CBOW in biomedical domain tasks \cite{Chiu2016}. Therefore, in future work, the effect of different \textit{word2vec} parameters on our ICD-9 code or category classifier will be further explored.

Further research on what words affect the probability of a prediction could improve our understanding of the relationship between symptoms and diagnosis. The probability observation could also change our preprocessing and feature extraction methods and ultimately improve our deep learning models. 

\section{Conclusion}
This study evaluates different NLP deep learning based models and feature extraction methods. It also establishes an empirical evaluation for learning-based automatic code assignment from the MIMIC-III discharge summary. The models are based on deep learning NLP frameworks that automatically assign clinical ICD-9 codes from free-text clinical notes. The deep learning models for predicting the top 10 ICD-9 codes and categories performed better than our baseline models that used traditional learning algorithms (best $F_1$ results: 0.6957 GRUs to 0.5320 Logistic Regression, and 0.7233 GRUs to 0.6313 FNNs, respectively). It was observed that the top 50 ICD-9 codes and categories results did not outperform our baseline ($F_1$ results: 0.3263 GRUs compared to 0.3662 Logistic Regression, and 0.3367 GRUs compare to 0.3651 FNNs). We believe that with more descriptive record collection and modern deep learning strategies, the predictive ability will likely increase. We also hope that our implementation and evaluation of the current state-of-the-art algorithms will serve as a baseline for further research on this topic.

\section*{Conflict of Interest Statement}
The authors declare that there is no conflict of interest. The authors do not have financial and personal relationships with other people or organizations that could inappropriately influence (bias) their work. 

\section*{Acknowledgement}
The authors would like to thank the anonymous reviewers for their valuable comments and feedback. This research did not receive any specific grant from funding agencies in the public, commercial, or not-for-profit sectors. The dataset used in this work is openly available developed by the MIT Lab for Computational Physiology.

\newpage

\bibliographystyle{unsrt}
\bibliography{reference}

\begin{thebibliography}{10}

\bibitem{black2011impact}
Ashly~D Black, Josip Car, Claudia Pagliari, Chantelle Anandan, Kathrin
  Cresswell, Tomislav Bokun, Brian McKinstry, Rob Procter, Azeem Majeed, and
  Aziz Sheikh.
\newblock The impact of ehealth on the quality and safety of health care: a
  systematic overview.
\newblock {\em PLoS Med}, 8(1):e1000387, 2011.

\bibitem{choi2016doctor}
Edward Choi, Mohammad~Taha Bahadori, Andy Schuetz, Walter~F Stewart, and Jimeng
  Sun.
\newblock Doctor ai: Predicting clinical events via recurrent neural networks.
\newblock In {\em Machine Learning for Healthcare Conference}, pages 301--318,
  2016.

\bibitem{martin2014big}
F~Martin-Sanchez, K~Verspoor, et~al.
\newblock Big data in medicine is driving big changes.
\newblock {\em Yearb Med Inform}, 9(1):14--20, 2014.

\bibitem{ferrao2013using}
Jose~C Ferrao, Filipe Janela, Monica~D Oliveira, and Henrique~MG Martins.
\newblock Using structured ehr data and svm to support icd-9-cm coding.
\newblock In {\em Healthcare Informatics (ICHI), 2013 IEEE International
  Conference on}, pages 511--516. IEEE, 2013.

\bibitem{pakhomov2006automating}
Serguei~VS Pakhomov, James~D Buntrock, and Christopher~G Chute.
\newblock Automating the assignment of diagnosis codes to patient encounters
  using example-based and machine learning techniques.
\newblock {\em Journal of the American Medical Informatics Association},
  13(5):516--525, 2006.

\bibitem{medori2010machine}
Julia Medori and C{\'e}drick Fairon.
\newblock Machine learning and features selection for semi-automatic icd-9-cm
  encoding.
\newblock In {\em Proceedings of the NAACL HLT 2010 Second Louhi Workshop on
  Text and Data Mining of Health Documents}, pages 84--89. Association for
  Computational Linguistics, 2010.

\bibitem{goldstein2007three}
Ira Goldstein, Anna Arzumtsyan, and {\"O}zlem Uzuner.
\newblock Three approaches to automatic assignment of icd-9-cm codes to
  radiology reports.
\newblock In {\em AMIA Annual Symposium Proceedings}, volume 2007, page 279.
  American Medical Informatics Association, 2007.

\bibitem{sutskever2014sequence}
Ilya Sutskever, Oriol Vinyals, and Quoc~V Le.
\newblock Sequence to sequence learning with neural networks.
\newblock In {\em Advances in neural information processing systems}, pages
  3104--3112, 2014.

\bibitem{collobert2011natural}
Ronan Collobert, Jason Weston, L{\'e}on Bottou, Michael Karlen, Koray
  Kavukcuoglu, and Pavel Kuksa.
\newblock Natural language processing (almost) from scratch.
\newblock {\em Journal of Machine Learning Research}, 12(Aug):2493--2537, 2011.

\bibitem{socher2011semi}
Richard Socher, Jeffrey Pennington, Eric~H Huang, Andrew~Y Ng, and
  Christopher~D Manning.
\newblock Semi-supervised recursive autoencoders for predicting sentiment
  distributions.
\newblock In {\em Proceedings of the conference on empirical methods in natural
  language processing}, pages 151--161. Association for Computational
  Linguistics, 2011.

\bibitem{johnson2016mimic}
Alistair~EW Johnson, Tom~J Pollard, Lu~Shen, H~Lehman Li-wei, Mengling Feng,
  Mohammad Ghassemi, Benjamin Moody, Peter Szolovits, Leo~Anthony Celi, and
  Roger~G Mark.
\newblock Mimic-iii, a freely accessible critical care database.
\newblock {\em Scientific data}, 3:160035, 2016.

\bibitem{larkey1995automatic}
Leah~S Larkey and W~Bruce Croft.
\newblock Automatic assignment of icd9 codes to discharge summaries.
\newblock Technical report, Technical report, University of Massachusetts at
  Amherst, Amherst, MA, 1995.

\bibitem{pestian2007shared}
John~P Pestian, Christopher Brew, Pawe{\l} Matykiewicz, Dj~J Hovermale, Neil
  Johnson, K~Bretonnel Cohen, and W{\l}odzis{\l}aw Duch.
\newblock A shared task involving multi-label classification of clinical free
  text.
\newblock In {\em Proceedings of the Workshop on BioNLP 2007: Biological,
  Translational, and Clinical Language Processing}, pages 97--104. Association
  for Computational Linguistics, 2007.

\bibitem{ribeiro2001experimental}
Berthier Ribeiro-Neto, Alberto~HF Laender, and Luciano~RS De~Lima.
\newblock An experimental study in automatically categorizing medical
  documents.
\newblock {\em Journal of the Association for Information Science and
  Technology}, 52(5):391--401, 2001.

\bibitem{meystre2008extracting}
St{\'e}phane~M Meystre, Guergana~K Savova, Karin~C Kipper-Schuler, John~F
  Hurdle, et~al.
\newblock Extracting information from textual documents in the electronic
  health record: a review of recent research.
\newblock {\em Yearb Med Inform}, 35(128):44, 2008.

\bibitem{ling2017methods}
Yuan Ling.
\newblock {\em Methods and Techniques for Clinical Text Modeling and
  Analytics}.
\newblock PhD thesis, Drexel University, 2017.

\bibitem{krizhevsky2012imagenet}
Alex Krizhevsky, Ilya Sutskever, and Geoffrey~E Hinton.
\newblock Imagenet classification with deep convolutional neural networks.
\newblock In {\em Advances in neural information processing systems}, pages
  1097--1105, 2012.

\bibitem{graves2013speech}
Alex Graves, Abdel-rahman Mohamed, and Geoffrey Hinton.
\newblock Speech recognition with deep recurrent neural networks.
\newblock In {\em Acoustics, speech and signal processing (icassp), 2013 ieee
  international conference on}, pages 6645--6649. IEEE, 2013.

\bibitem{choi2016using}
Edward Choi, Andy Schuetz, Walter~F Stewart, and Jimeng Sun.
\newblock Using recurrent neural network models for early detection of heart
  failure onset.
\newblock {\em Journal of the American Medical Informatics Association}, page
  ocw112, 2016.

\bibitem{lipton2015learning}
Zachary~C Lipton, David~C Kale, Charles Elkan, and Randall Wetzell.
\newblock Learning to diagnose with lstm recurrent neural networks.
\newblock {\em arXiv preprint arXiv:1511.03677}, 2015.

\bibitem{choi2016retain}
Edward Choi, Mohammad~Taha Bahadori, Jimeng Sun, Joshua Kulas, Andy Schuetz,
  and Walter Stewart.
\newblock Retain: An interpretable predictive model for healthcare using
  reverse time attention mechanism.
\newblock In {\em Advances in Neural Information Processing Systems}, pages
  3504--3512, 2016.

\bibitem{zeng2019automatic}
Min Zeng, Min Li, Zhihui Fei, Ying Yu, Yi~Pan, and Jianxin Wang.
\newblock Automatic icd-9 coding via deep transfer learning.
\newblock {\em Neurocomputing}, 324:43--50, 2019.

\bibitem{prakash2017condensed}
Aaditya Prakash, Siyuan Zhao, Sadid~A Hasan, Vivek~V Datla, Kathy Lee, Ashequl
  Qadir, Joey Liu, and Oladimeji Farri.
\newblock Condensed memory networks for clinical diagnostic inferencing.
\newblock In {\em AAAI}, pages 3274--3280, 2017.

\bibitem{shickel2018deep}
Benjamin Shickel, Patrick~James Tighe, Azra Bihorac, and Parisa Rashidi.
\newblock Deep ehr: a survey of recent advances in deep learning techniques for
  electronic health record (ehr) analysis.
\newblock {\em IEEE journal of biomedical and health informatics},
  22(5):1589--1604, 2018.

\bibitem{mikolov2013distributed}
Tomas Mikolov, Ilya Sutskever, Kai Chen, Greg~S Corrado, and Jeff Dean.
\newblock Distributed representations of words and phrases and their
  compositionality.
\newblock In {\em Advances in neural information processing systems}, pages
  3111--3119, 2013.

\bibitem{Chiu2016}
Billy Chiu, Gamal Crichton, Anna Korhonen, and Sampo Pyysalo.
\newblock How to train good word embeddings for biomedical nlp.
\newblock {\em Proceedings of the 15th Workshop on Biomedical Natural Language
  Processing}, pages 166--174, 2016.

\bibitem{tsoumakas2007multi}
Grigorios Tsoumakas and Ioannis Katakis.
\newblock Multi-label classification: An overview.
\newblock {\em International Journal of Data Warehousing and Mining (IJDWM)},
  3(3):1--13, 2007.

\bibitem{faruqui2015retrofitting}
Manaal Faruqui, Jesse Dodge, Sujay~Kumar Jauhar, Chris Dyer, Eduard Hovy, and
  Noah~A Smith.
\newblock Retrofitting word vectors to semantic lexicons.
\newblock In {\em Proceedings of the 2015 Conference of the North American
  Chapter of the Association for Computational Linguistics: Human Language
  Technologies}, pages 1606--1615, 2015.

\bibitem{kenter2015ad}
Tom Kenter, Melvin Wevers, Pim Huijnen, and Maarten de~Rijke.
\newblock Ad hoc monitoring of vocabulary shifts over time.
\newblock In {\em Proceedings of the 24th ACM International on Conference on
  Information and Knowledge Management}, pages 1191--1200. ACM, 2015.

\bibitem{yih2011learning}
Wen-tau Yih, Kristina Toutanova, John~C Platt, and Christopher Meek.
\newblock Learning discriminative projections for text similarity measures.
\newblock In {\em Proceedings of the Fifteenth Conference on Computational
  Natural Language Learning}, pages 247--256. Association for Computational
  Linguistics, 2011.

\bibitem{shen2014learning}
Yelong Shen, Xiaodong He, Jianfeng Gao, Li~Deng, and Gr{\'e}goire Mesnil.
\newblock Learning semantic representations using convolutional neural networks
  for web search.
\newblock In {\em Proceedings of the 23rd International Conference on World
  Wide Web}, pages 373--374. ACM, 2014.

\bibitem{kim2014convolutional}
Yoon Kim.
\newblock Convolutional neural networks for sentence classification.
\newblock In {\em Proceedings of the 2014 Conference on Empirical Methods in
  Natural Language Processing (EMNLP)}, pages 1746--1751, 2014.

\bibitem{graves2012supervised}
Alex Graves et~al.
\newblock {\em Supervised sequence labelling with recurrent neural networks},
  volume 385.
\newblock Springer, 2012.

\bibitem{bengio1994learning}
Yoshua Bengio, Patrice Simard, and Paolo Frasconi.
\newblock Learning long-term dependencies with gradient descent is difficult.
\newblock {\em IEEE transactions on neural networks}, 5(2):157--166, 1994.

\bibitem{hochreiter1997long}
Sepp Hochreiter and J{\"u}rgen Schmidhuber.
\newblock Long short-term memory.
\newblock {\em Neural computation}, 9(8):1735--1780, 1997.

\bibitem{cho2014properties}
Kyunghyun Cho, Bart van Merri{\"e}nboer, Dzmitry Bahdanau, and Yoshua Bengio.
\newblock On the properties of neural machine translation: Encoder--decoder
  approaches.
\newblock {\em Syntax, Semantics and Structure in Statistical Translation},
  page 103, 2014.

\bibitem{zhang2014review}
Min-Ling Zhang and Zhi-Hua Zhou.
\newblock A review on multi-label learning algorithms.
\newblock {\em IEEE transactions on knowledge and data engineering},
  26(8):1819--1837, 2014.

\bibitem{agrawal2013multi}
Rahul Agrawal, Archit Gupta, Yashoteja Prabhu, and Manik Varma.
\newblock Multi-label learning with millions of labels: Recommending advertiser
  bid phrases for web pages.
\newblock In {\em Proceedings of the 22nd international conference on World
  Wide Web}, pages 13--24. ACM, 2013.

\bibitem{dembczynski2012label}
Krzysztof Dembczy{\'n}ski, Willem Waegeman, Weiwei Cheng, and Eyke
  H{\"u}llermeier.
\newblock On label dependence and loss minimization in multi-label
  classification.
\newblock {\em Machine Learning}, 88(1-2):5--45, 2012.

\bibitem{lea2017temporal}
Colin Lea, Michael~D Flynn, Rene Vidal, Austin Reiter, and Gregory~D Hager.
\newblock Temporal convolutional networks for action segmentation and
  detection.
\newblock In {\em proceedings of the IEEE Conference on Computer Vision and
  Pattern Recognition}, pages 156--165, 2017.

\bibitem{BaiTCN2018}
Shaojie Bai, J.~Zico Kolter, and Vladlen Koltun.
\newblock An empirical evaluation of generic convolutional and recurrent
  networks for sequence modeling.
\newblock {\em arXiv:1803.01271}, 2018.

\bibitem{le2014distributed}
Quoc Le and Tomas Mikolov.
\newblock Distributed representations of sentences and documents.
\newblock In {\em International Conference on Machine Learning}, pages
  1188--1196, 2014.

\end{thebibliography}

\newpage
\begin{appendices}

\section{Model Performance for Top 10 Label Codes}

\begin{table}[ht]
\centering
\resizebox{\columnwidth}{!}{%
\begin{tabu}{|c|c c c c|c c c c|}
\hline
 & \multicolumn{4}{|c|}{Training} & \multicolumn{4}{|c|}{Test} \\ \hline
Model & Precision & Recall & Accuracy & $F_1$ & Precision & Recall & Accuracy & $F_1$ \\ \hline
Logistic Regression & 0.9564 & 0.9440 & 0.9786 & 0.9501 & 0.5801 & 0.4934 & 0.8392 & 0.5320	 \\ \hline
Random Forests & 0.9989 & 0.6988 & 0.9501 & 0.8086 & 0.7573 & 0.2340 & 0.8432 & 0.3219	  \\ \hline
\rowfont{\color{red}}Feed-forward NN & 0.7933 & 0.5742 & 0.8998 & 0.6457 & 0.6810 & 0.4634 & 0.8622 & 0.5323 \\ \hline 
CNNs & 0.8312 & 0.6713 & 0.9165 & 0.7371 & 0.7408 & 0.5687 & 0.8832 & 0.6373 \\ \hline
LSTM RNNs & 0.8106 & 0.6971 & 0.9154 & 0.7445 & 0.7574 & 0.6380 & 0.8950 & 0.6874 \\ \hline
GRU RNNs & 0.7936 &	0.6971 & 0.9126	& 0.7397 & 0.7502 & 0.6519 & 0.8967	& 0.6957 \\ \hline
\end{tabu}
}
\caption{Model Performance for \textit{top-10-code} \label{tab:model-perf-top-10-code} }
\end{table}

\section{Model Performance for Top 50 Codes}

\renewcommand{\thefootnote}{\alph{footnote}}

\begin{table}[h]
\centering
\resizebox{\columnwidth}{!}{%
\begin{tabu}{|c|c c c c|c c c c|}
\hline
 & \multicolumn{4}{|c|}{Training} & \multicolumn{4}{|c|}{Test} \\ \hline
Model & Precision & Recall & Accuracy & $F_1$ & Precision & Recall & Accuracy & $F_1$ \\ \hline
Logistic Regression & 0.9863 & 0.9768 & 0.9945 & 0.9815 & 0.4372 & 0.3213 & 0.9148 & 0.3662 \\ \hline
Random Forests & 0.9985 & 0.2852 & 0.9451 & 0.3866 & 0.5377 & 0.0953 & 0.9220 & 0.1155	  \\ \hline
\rowfont{\color{red}}Feed-forward NN & 0.2490 & 0.1138 & 0.9224 & 0.1268 & 0.2251 & 0.1090 & 0.9212 & 0.1215 \\ \hline
CNNs & 0.6085 \footnotemark & 0.2663 & 0.9365 & 0.3200 \footnotemark[\value{footnote}] & 0.4792 \footnotemark[\value{footnote}] & 0.2169 & 0.9286 & 0.2609 \footnotemark[\value{footnote}] \\ \hline
LSTM RNNs & 0.3526 \footnotemark[\value{footnote}] & 0.1642 & 0.9325 & 0.1891 \footnotemark[\value{footnote}] & 0.4022 \footnotemark[\value{footnote}] & 0.1445 & 0.9286 & 0.1659 \footnotemark[\value{footnote}] \\ \hline
GRU RNNs & 0.6539 \footnotemark[\value{footnote}] & 0.3433 & 0.9460 & 0.3947 \footnotemark[\value{footnote}] & 0.5592 \footnotemark[\value{footnote}] & 0.2782 & 0.9354 & 0.3263 \footnotemark[\value{footnote}] \\ \hline
\end{tabu}
}
\caption{Model Performance for \textit{top-50-code}  \label{tab:model-perf-top-50-code} }
\end{table}


\begin{table}[ht]
\centering
\resizebox{\columnwidth}{!}{%
\begin{tabu}{|c|c c c c|c c c c|}
\hline
 & \multicolumn{4}{|c|}{Training} & \multicolumn{4}{|c|}{Test} \\ \hline
Model & Precision & Recall & Accuracy & $F_1$ & Precision & Recall & Accuracy & $F_1$ \\ \hline
Logistic Regression & 0.9564 & 0.9440 & 0.9786 & 0.9501 & 0.5801 & 0.4934 & 0.8392 & 0.5320 \\ \hline
Random Forests & 0.9946 & 0.4937 & 0.9110 & 0.6305 & 0.7869 & 0.2009 & 0.8395 & 0.2822 \\ \hline
\rowfont{\color{red}}Feed-forward NN & 0.5266\footnotemark[\value{footnote}] & 0.2783 & 0.8408 & 0.3380\footnotemark[\value{footnote}] & 0.5143 \footnotemark[\value{footnote}] & 0.2676 & 0.8370 & 0.3276 \footnotemark[\value{footnote}] \\ \hline
CNNs & 0.7708 & 0.4673 & 0.8858 & 0.5377 & 0.6784 & 0.4109 & 0.8650 & 0.4739 \\ \hline
LSTM RNNs & 0.6204\footnotemark[\value{footnote}] & 0.3829 & 0.8805 & 0.4348\footnotemark[\value{footnote}]  & 0.5748 & 0.3526 & 0.8688 & 0.4025\footnotemark[\value{footnote}]  \\ \hline
GRU RNNs & 0.8351 & 0.6474 & 0.9168 & 0.7181 & 0.7520 & 0.5618 & 0.8871 & 0.6328 \\ \hline
\end{tabu}
}
\caption{Model Performance for \textit{top-50-code} (first 10)  \label{tab:model-perf-top-50-code-first10} }
\end{table}

\footnotetext{result contained $nan$. Computed by replacing $nan$ with zero.}

\newpage
\section{Model Performance for Top 10 Label Categories}
\begin{table}[ht]
\centering
\resizebox{\columnwidth}{!}{%
\begin{tabu}{|c|c c c c|c c c c|}
\hline
 & \multicolumn{4}{|c|}{Training} & \multicolumn{4}{|c|}{Test} \\ \hline
Model & Precision & Recall & Accuracy  & $F_1$ & Precision & Recall & Accuracy & $F_1$ \\ \hline
Logistic Regression & 0.9437 & 0.9309 & 0.9652 & 0.9372 & 0.6458 & 0.5856 & 0.7994 & 0.6141  \\ \hline
Random Forests & 0.9983 & 0.8134 & 0.9500 & 0.8954 & 0.7653 & 0.3801 & 0.8019 & 0.4966	  \\ \hline
\rowfont{\color{red}}Feed-forward NN & 0.7989 & 0.6456 & 0.8632 & 0.7083 & 0.7334 & 0.5633 & 0.8272 & 0.6314 \\ \hline
CNNs & 0.8039 & 0.6637 & 0.8681 & 0.7128	& 0.7613 & 0.6126 & 0.8446 & 0.6657 \\ \hline
LSTM RNNs & 0.8146 & 0.6807 & 0.8749 & 0.7343 & 0.7926 & 0.6536 & 0.8622	 & 0.7090 \\ \hline
GRU RNNs & 0.8150 & 0.7613 & 0.8909 & 0.7861 & 0.7580 & 0.6941 & 0.8588 & 0.7233 \\ \hline
\end{tabu}
}
\caption{Model Performance for \textit{top-10-cat} \label{tab:model-perf-top-10-cat} }
\end{table}

\section{Model Performance for Top 50 Label Categories}
\begin{table}[ht]
\centering
\resizebox{\columnwidth}{!}{%
\begin{tabu}{|c|c c c c|c c c c|}
\hline
 & \multicolumn{4}{|c|}{Training} & \multicolumn{4}{|c|}{Test} \\ \hline
Model & Precision & Recall & Accuracy  & $F_1$ & Precision & Recall & Accuracy & $F_1$ \\ \hline
Logistic Regression & 0.9750 & 0.9572 & 0.9887 & 0.9659 & 0.4858 & 0.3894 & 0.8841 & 0.4301	  \\ \hline
Random Forests &0.9986 & 0.3294 & 0.9277 & 0.4465 & 0.6568 & 0.1142 & 0.8906 & 0.1576  \\ \hline
\rowfont{\color{red}}Feed-forward NN & 0.3522 & 0.1654 & 0.8940 & 0.2007 & 0.3600 & 0.1557 & 0.8909 & 0.1901 \\ \hline
CNNs & 0.7428 & 0.3262 & 0.9163 & 0.3870 & 0.5635 \footnotemark[\value{footnote}] & 0.2770 & 0.9035 & 0.3301 \footnotemark[\value{footnote}] \\ \hline
LSTM RNNs & 0.7117 \footnotemark[\value{footnote}] & 0.3363 & 0.9194 & 0.3804 \footnotemark[\value{footnote}] & 0.5869 & 0.2945 & 0.9087 & 0.3367 \footnotemark[\value{footnote}] \\ \hline
GRU RNNs & 0.6695 \footnotemark[\value{footnote}] & 0.3227 & 0.9179 & 0.3726 \footnotemark[\value{footnote}] & 0.5611 \footnotemark[\value{footnote}] & 0.2809 & 0.9067 & 0.3266 \footnotemark[\value{footnote}] \\ \hline
\end{tabu}
}
\caption{Model Performance for \textit{top-50-cat}  \label{tab:model-perf-top-50-cat} }
\end{table}


\begin{table}[ht]
\centering
\resizebox{\columnwidth}{!}{%
\begin{tabu}{|c|c c c c|c c c c|}
\hline
 & \multicolumn{4}{|c|}{Training} & \multicolumn{4}{|c|}{Test} \\ \hline
Model & Precision & Recall & Accuracy & $F_1$ & Precision & Recall & Accuracy & $F_1$ \\ \hline
Logistic Regression & 0.9437 & 0.9309 & 0.9652 & 0.9372 & 0.6458 & 0.5856 & 0.7994 & 0.6141 \\ \hline
Random Forests & 0.9937 & 0.6321 & 0.8999 & 0.7687 & 0.7877 & 0.3282 & 0.7944 & 0.4512 \\ \hline
\rowfont{\color{red}}Feed-forward NN & 0.6905 & 0.4762 & 0.8043 & 0.5535 & 0.6795 & 0.4562 & 0.7955 & 0.5347 \\ \hline
CNNs & 0.7945 & 0.6652 & 0.8670 & 0.7142 & 0.7296 & 0.5979 & 0.8345 & 0.6481 \\ \hline
LSTM RNNs & 0.7963 & 0.6863 & 0.8768 & 0.7213 & 0.7515 & 0.6362 & 0.8514 & 0.6738 \\ \hline
GRU RNNs & 0.7901 & 0.6803 & 0.8729 & 0.7213 & 0.7382 & 0.6196 & 0.8442 & 0.6641 \\ \hline
\end{tabu}
}
\caption{Model Performance for \textit{top-50-cat} (first 10)  \label{tab:model-perf-top-50-cat-first10} }
\end{table}

\footnotetext{result contained $nan$. Computed by replacing $nan$ with zero.}







\newpage
\section{Best Performance Models (LSTMs, GRUs, CNNs) Precision-Recall Curve}
\label{sec:precision_recall_curve}

\begin{figure}[htbp]
\centering
\subfigure[LSTMs Top 10 Codes]{\includegraphics[width=0.48\linewidth]{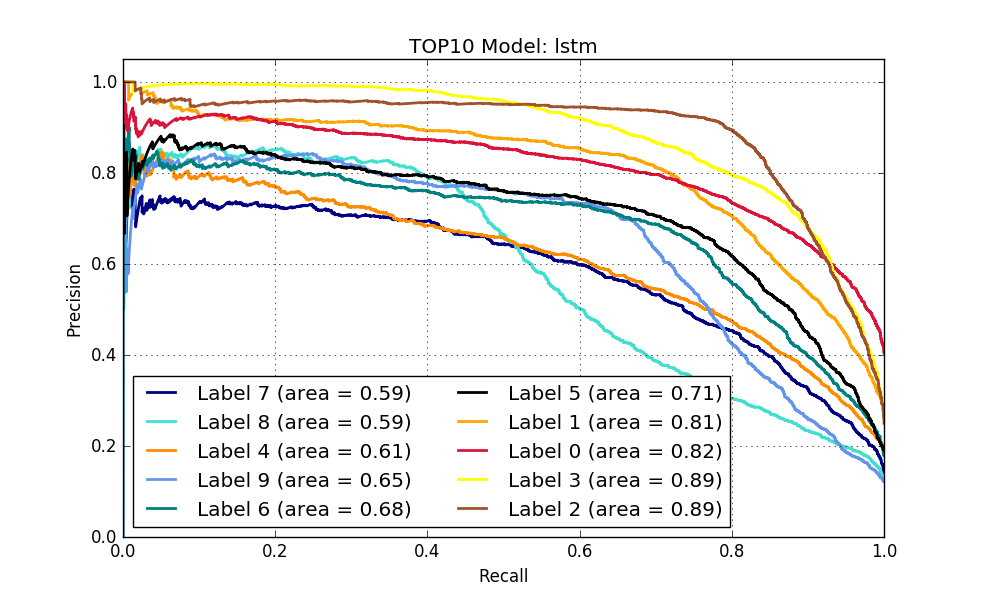}\label{fig:lstm_10}}
\subfigure[LSTMs Top 10 Categories]{\includegraphics[width=0.48\linewidth]{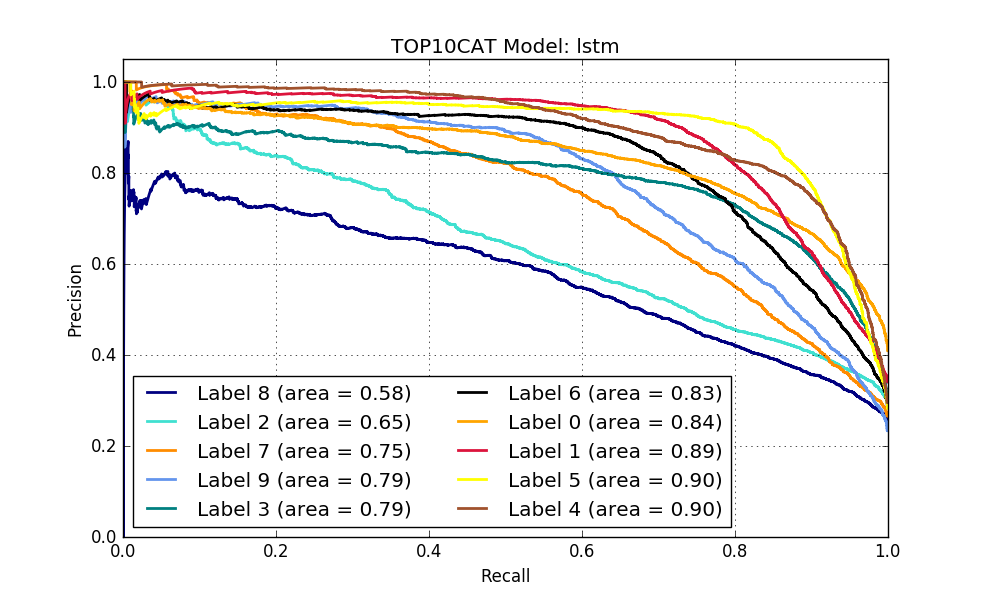}\label{fig:lstm_10cat}}

\subfigure[LSTMs Top 50 Codes]{\includegraphics[width=0.48\linewidth]{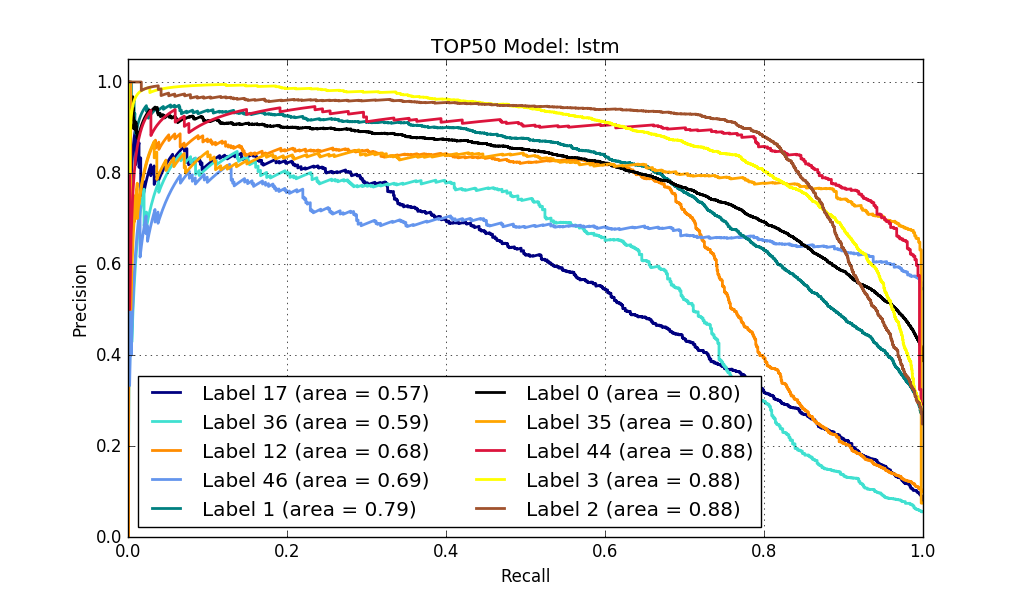}\label{fig:lstm_50}}
\subfigure[LSTMs Top 50 Categories]{\includegraphics[width=0.48\linewidth]{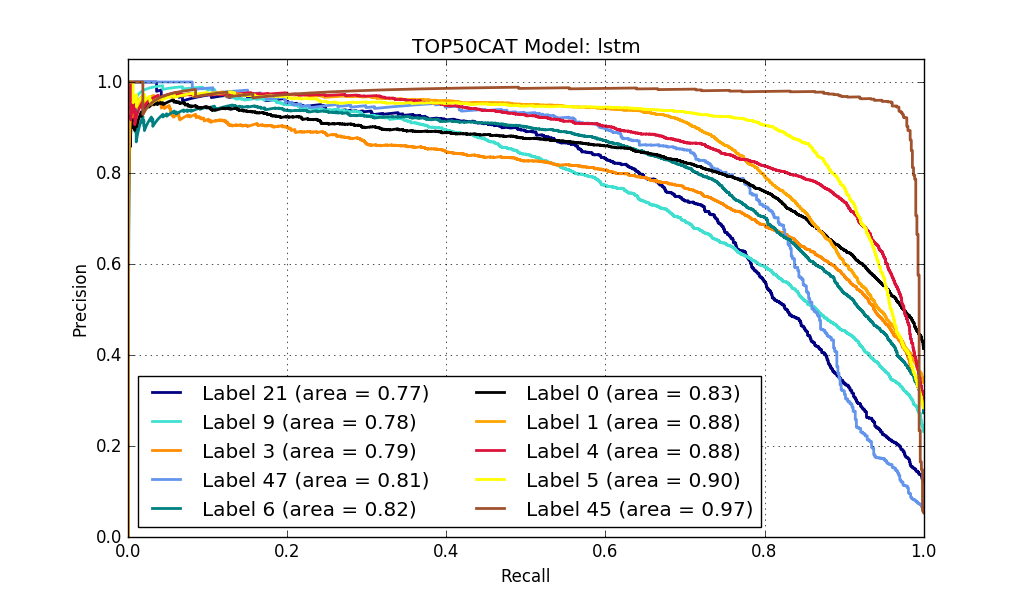}\label{fig:lstm_50cat}}

\subfigure[GRUs Top 10 Codes]{\includegraphics[width=0.48\linewidth]{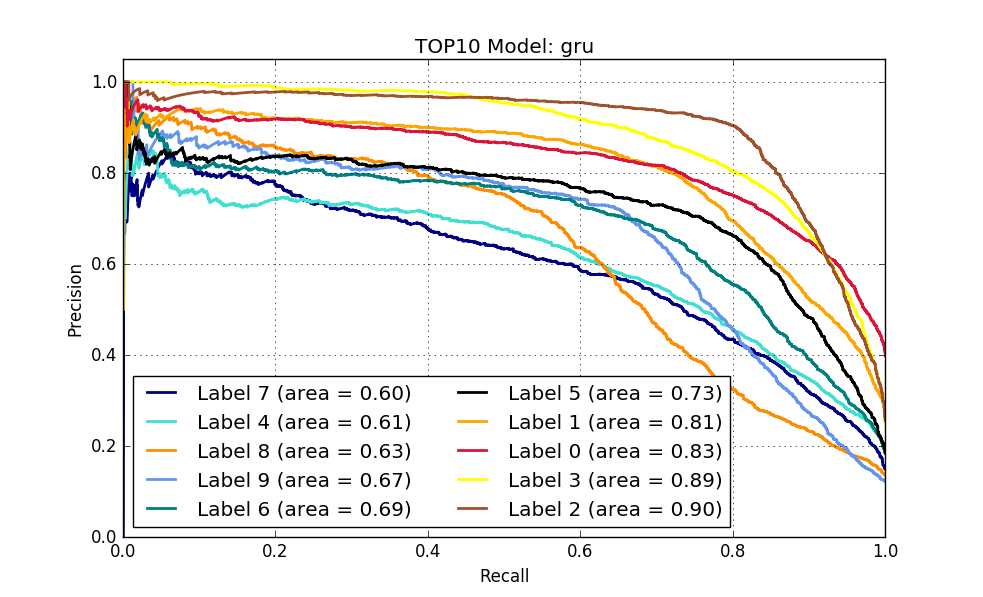}\label{fig:gru_10}}
\subfigure[GRUs Top 10 Categories]{\includegraphics[width=0.48\linewidth]{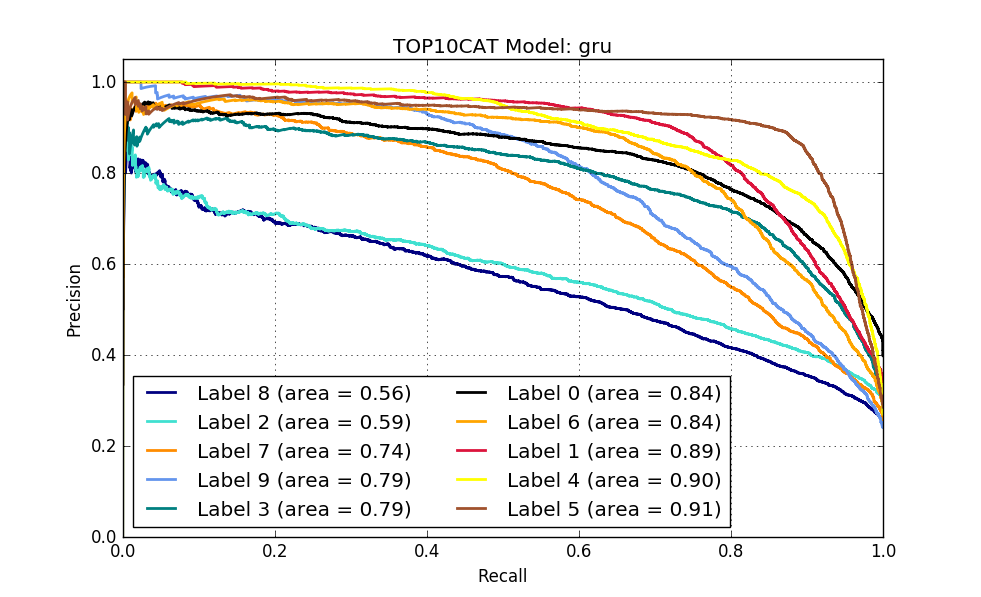}\label{fig:gru_10cat}}
\end{figure}


\begin{figure}[htbp]
\centering
\subfigure[GRUs Top 50 Codes]{\includegraphics[width=0.48\linewidth]{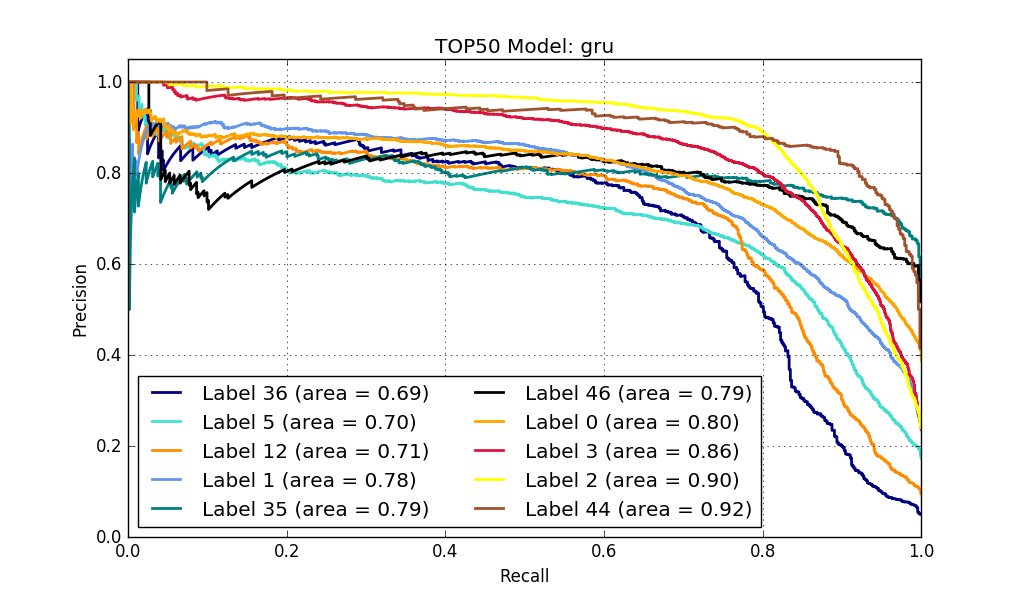}\label{fig:gru_50}}
\subfigure[GRUs Top 50 Categories]{\includegraphics[width=0.48\linewidth]{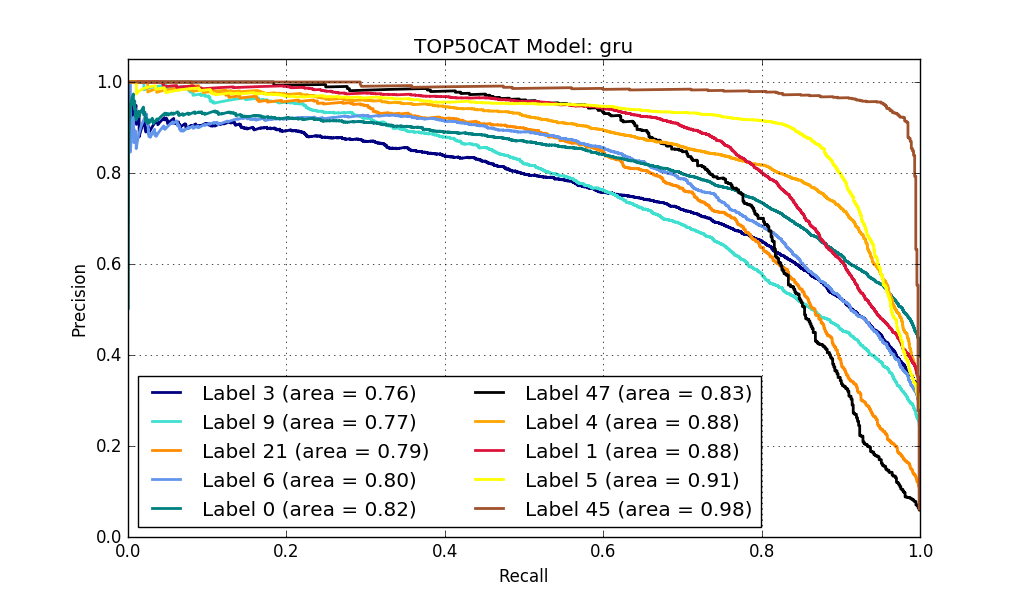}\label{fig:gru_50cat}}


\subfigure[CNNs Top 10 Codes]{\includegraphics[width=0.48\linewidth]{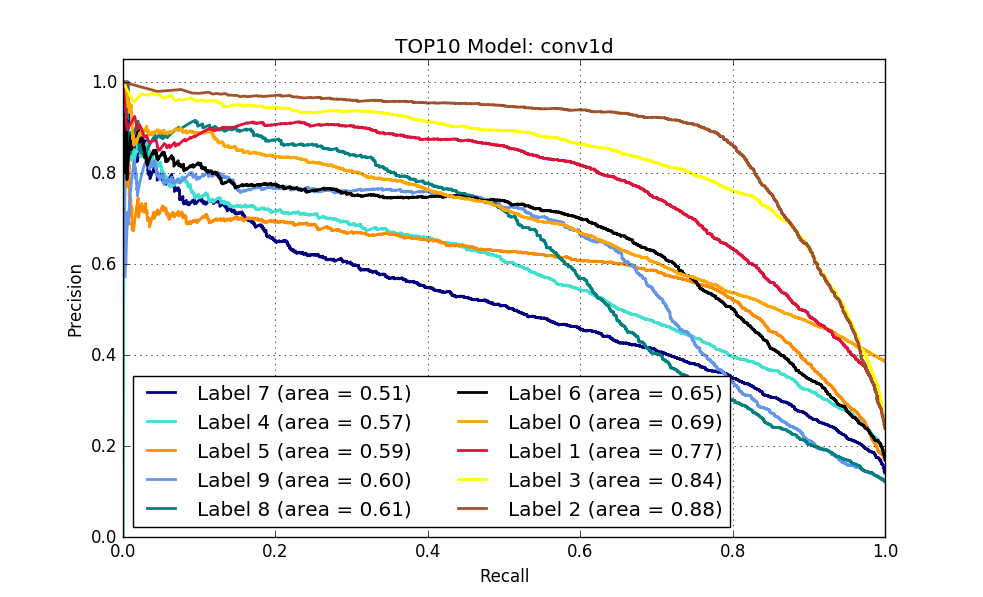}\label{fig:conv1d_10}}
\subfigure[CNNs Top 10 Categories]{\includegraphics[width=0.48\linewidth]{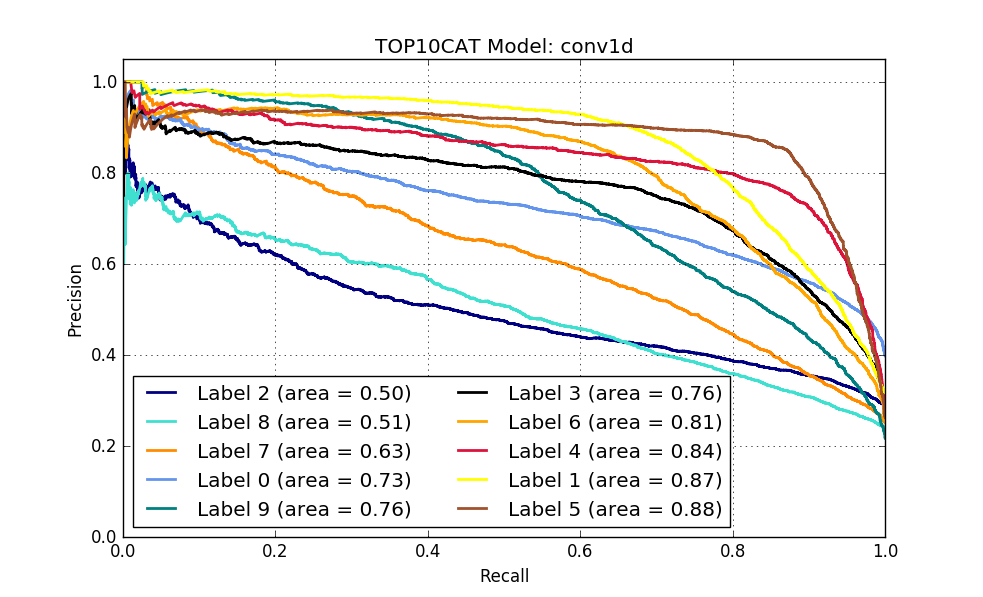}\label{fig:conv1d_10cat}}
\subfigure[CNNs Top 50 Codes]{\includegraphics[width=0.48\linewidth]{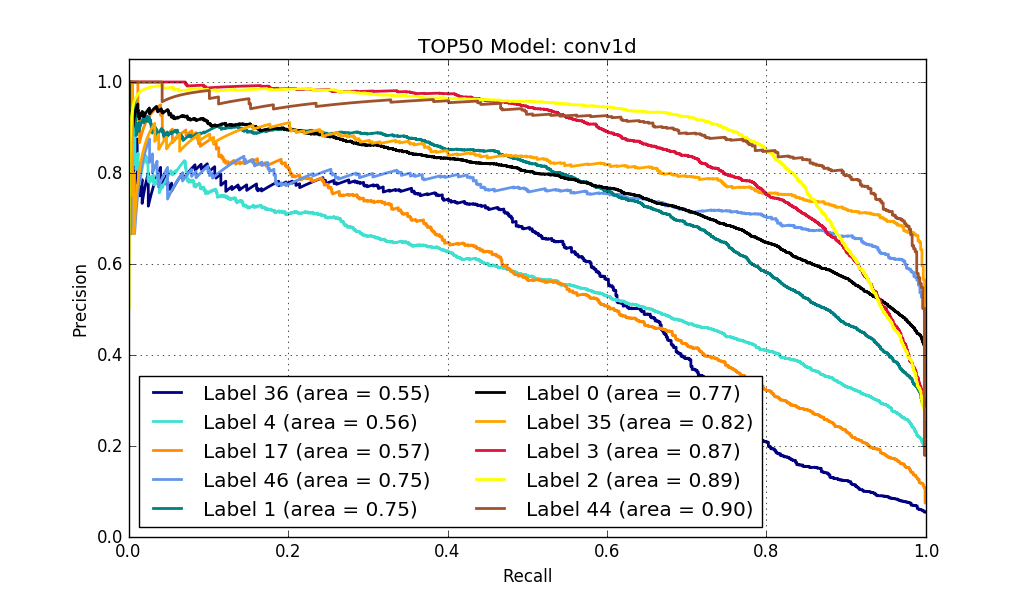}\label{fig:conv1d_50}}
\subfigure[CNNs Top 50 Categories]{\includegraphics[width=0.48\linewidth]{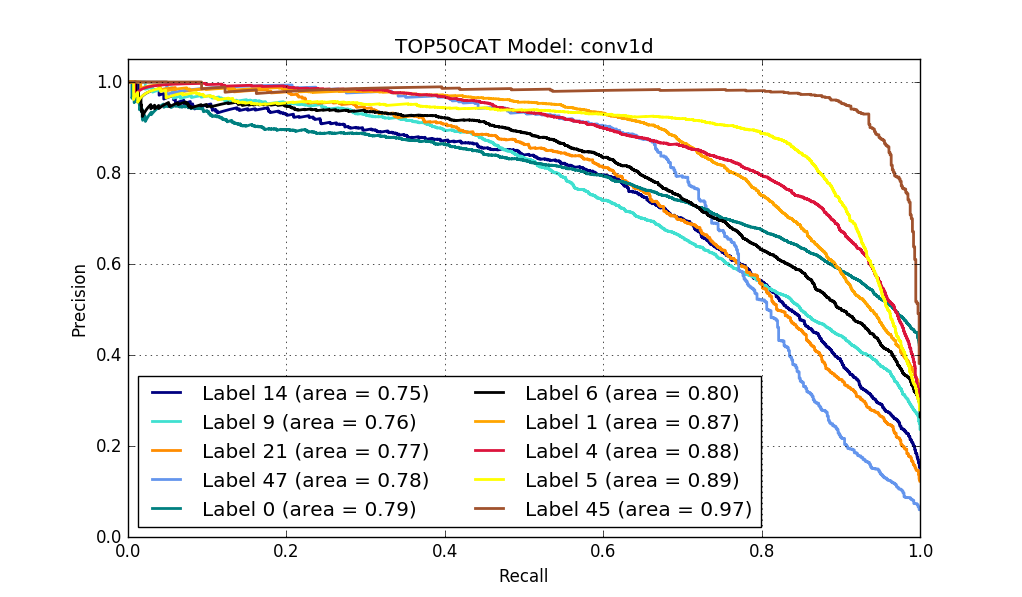}\label{fig:conv1d_50cat}}
\end{figure}

\end{appendices}

\end{document}